\pdfoutput=1

\documentclass[11pt]{article}

\usepackage{ACL2023}

\usepackage{times}
\usepackage{latexsym}

\usepackage[T1]{fontenc}

\usepackage[utf8]{inputenc}

\usepackage{microtype}

\usepackage{inconsolata}

\usepackage{times}
\usepackage{latexsym}

\usepackage[T1]{fontenc}

\usepackage[utf8]{inputenc}

\usepackage{microtype}

\usepackage{inconsolata}

\usepackage{makecell}
\usepackage{multicol}
\usepackage{caption}
\usepackage{subcaption}
\usepackage{appendix}
\usepackage{url}

\usepackage{xspace}
\usepackage{mathtools}
\usepackage{amssymb}
\usepackage{pifont}
\usepackage{tablefootnote}
\usepackage{color, colortbl}
\usepackage{xstring}
\usepackage{comment}
\usepackage{float}
\restylefloat{table}
\usepackage{array,booktabs,makecell}
\usepackage[normalem]{ulem}
\usepackage{arydshln}
\usepackage{titlesec}
 \usepackage{color,soul}

\usepackage{placeins}
\usepackage{wrapfig,lipsum,booktabs}
\usepackage{array}
\usepackage{microtype}
\usepackage{linguex}
\usepackage{tipa}
\usepackage[british]{babel}
\usepackage{color,soul}
\usepackage{arydshln}
\usepackage{adjustbox}
\usepackage[utf8]{inputenc}
\usepackage{placeins}
\usepackage{hyperref}
\usepackage{expex}
\usepackage{tabto}
\usepackage{gb4e}
\noautomath
\usepackage{enumitem}
\usepackage{epstopdf}
\usepackage{hhline}
\usepackage{qtree} 
\usepackage{times}

\usepackage{amsmath}
\usepackage{cleveref}

\usepackage{graphicx}
\usepackage{booktabs}
\usepackage{multirow}

\newcommand\blfootnote[1]{%
  \begingroup
  \renewcommand\thefootnote{}\footnote{#1}%
  \addtocounter{footnote}{-1}%
  \endgroup
}
\def \ourmodelnlg{Cheetah}

\title{Cheetah\includegraphics[scale=0.05]{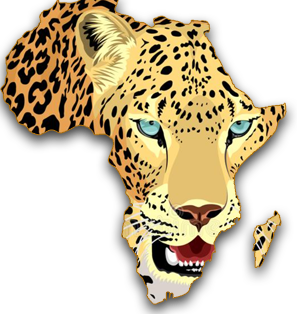}: Natural Language Generation for 517 African Languages}

\author{\normalsize Ife Adebara$^{1,\star}$ ~ AbdelRahim Elmadany$^{1,\star}$ ~ Muhammad Abdul-Mageed$^{1,2}$ \\
\normalsize $^{1}$Deep Learning \& Natural Language Processing Group,
  The University of British Columbia\\\normalsize  $^{2}$Department of Natural Language Processing \& Department of Machine Learning, MBZUAI\\ %
  \texttt{\normalsize \{ife.adebara@,a.elmadany@,muhammad.mageed@\}ubc.ca}}

\begin{document}
\maketitle
\begin{abstract}
Low-resource African languages pose unique challenges for natural language processing (NLP) tasks, including natural language generation (NLG). In this paper, we develop \ourmodelnlg, a massively multilingual NLG language model for African languages. \ourmodelnlg~supports $517$ African languages and language varieties, allowing us to address the scarcity of NLG resources and provide a solution to foster linguistic diversity. We demonstrate the effectiveness of~\ourmodelnlg~through comprehensive evaluations across six generation downstream tasks. In five of the six tasks,~\ourmodelnlg~significantly outperforms other models, showcasing its remarkable performance for generating coherent and contextually appropriate text in a wide range of African languages. We additionally conduct a detailed human evaluation to delve deeper into the linguistic capabilities of~\ourmodelnlg. The introduction of \ourmodelnlg~has far-reaching benefits for linguistic diversity. By leveraging pretrained models and adapting them to specific languages, our approach facilitates the development of practical NLG applications for African communities. The findings of this study contribute to advancing NLP research in low-resource settings, enabling greater accessibility and inclusion for African languages in a rapidly expanding digital landscape.  We will publicly release our models for research. \footnote{\href{https://github.com/UBC-NLP/Cheetah}{https://github.com/UBC-NLP/Cheetah}}\blfootnote{ $^{\star}$ Authors contributed equally.}\\
\end{abstract}

\section{Introduction}\label{sec:intro}
The linguistic diversity present in African languages poses unique challenges for NLG systems. With over $2,000$ languages spoken across the African continent \cite{ethnologue}, the need for effective NLG solutions that can accommodate this rich linguistic ecosystem cannot be over-emphasized. This is especially important because traditional NLG approaches have primarily focused on high-resource languages, such as English and French due to the availability of large-scale datasets and resources. Consequently, low-resource languages, including numerous African languages, have been marginalized in NLG research and development. Developing robust NLG systems for the diverse needs of African communities is challenging due to the scarcity of extensive language datasets, limited linguistic research, and variations across these languages.
\begin{figure}[t]
  \centering
  \includegraphics[width=\columnwidth]{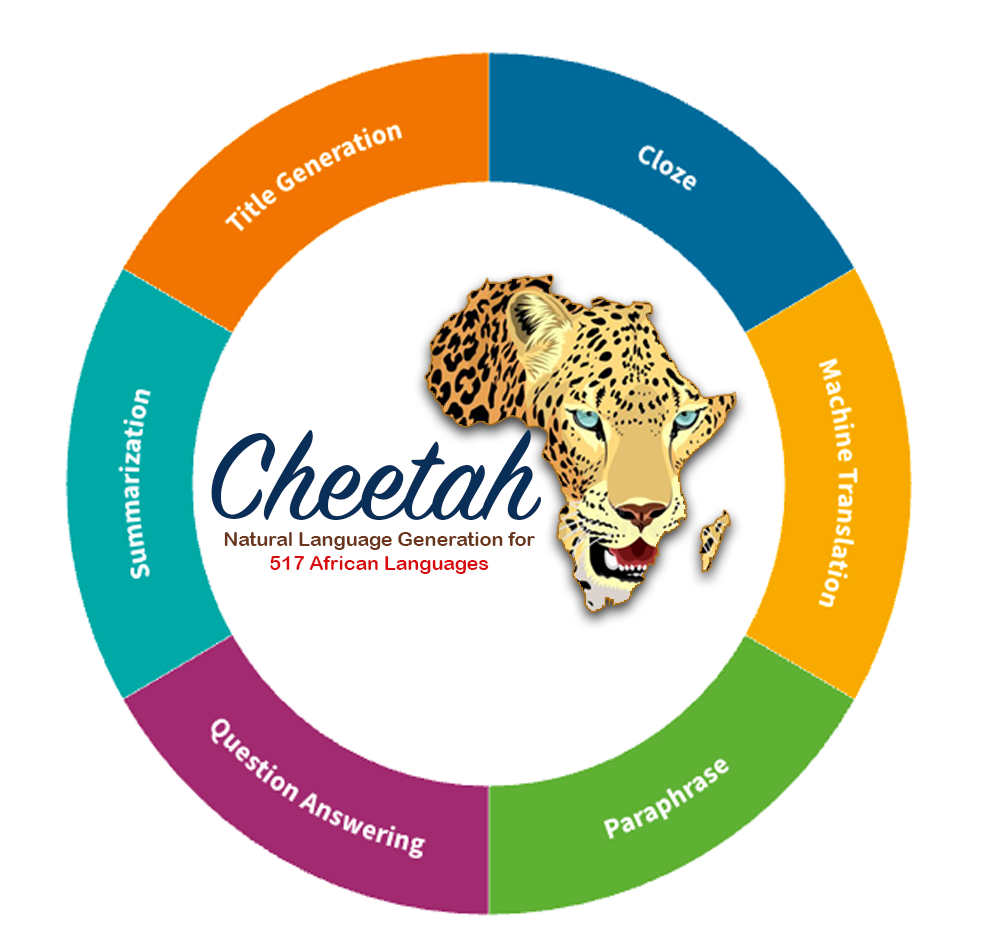}
\caption{ Cheetah is trained on $517$ African languages and language varieties across $14$ language families. The languages are domiciled in $50$ of $54$ African countries and are written in six different scripts.}
\label{fig:countries} 
\end{figure}
To address these challenges, recent advancements in language modeling and transfer learning techniques have shown promise in supporting NLG in low-resource languages. Pretrained language models, such as GPT-3 \cite{radford2018improving, radford2019language, brown_2020}, mT5 \cite{xue-etal-2021-mt5}, and mT0 \cite{muennighoff2022crosslingual}, have demonstrated remarkable capabilities in understanding and generating human-like text. These models capture the statistical regularities and syntactic structures of the languages they are trained on, making them valuable starting points for supporting NLG in low-resource settings.

In this paper, we present a pioneering work on NLG in African languages by introducing \ourmodelnlg: a novel language model (LM) specifically designed to support $517$ African languages and language varieties. To the best of our knowledge, \ourmodelnlg~supports the largest number of African languages and language varieties. Leveraging a vast corpus of text data collected from diverse sources, \ourmodelnlg~learns some intricate linguistic information that characterize each African language. The contributions of this research are three fold. \textbf{First}, we address the scarcity of NLG resources for African languages by providing a comprehensive language model that covers a wide range of languages spoken on the continent. \textbf{Second}, we demonstrate the efficacy of our approach through extensive evaluations across six downstream task clusters. Each cluster includes multiple languages, showcasing the model's ability to generate coherent and contextually appropriate text in different African languages. \textbf{Third}, we perform fine grained human analysis of \ourmodelnlg~using a controlled machine translation (MT) test set. This uncovers model behaviour that is not visible with automatic metrics.
By supporting NLG in African languages, we foster linguistic diversity, empower African communities to express themselves in their native languages, and bridge the digital divide. This paper serves as a foundational step towards promoting Afrocentric NLP~\cite{adebara-abdul-mageed-2022-towards} that prioritizes inclusivity and cultural preservation in language technology, emphasizing the importance of catering to the unique linguistic needs of diverse populations. 

The rest of the paper is organized as follows: In Section \ref{sec:litreview}, we discuss related work. In Section, \ref{sec:benchmark} we describe AfroNLG, the benchmark we create for evaluation. We provide details of~\ourmodelnlg~in Section~\ref{sec:afrot5}. We present performance of \ourmodelnlg~in Section \ref{sec:evaluation} and compare it to other multilingual models. We present controlled test sets in Section \ref{sec:cfg}. We conclude in Section \ref{sec:conc}, and outline a number of limitations and use cases for our work in Section \ref{sec:limits} and Section \ref{sec:ethics}.

 \section{Literature Review}\label{sec:litreview}

One of the challenges in NLG is to generate coherent and semantically meaningful text. Various approaches have been proposed, including template-based \cite{becker-2002-practical, deemter_2005}, rule-based \cite{dusek-jurcicek-2015-training, van-miltenburg-etal-2020-evaluation}, and statistical approaches \cite{li-etal-2016-statistics}. More recently, deep learning approaches \cite{sutskever_2014} including the transformer model  \cite{vaswani_2017} have achieved SoTA results in various NLG tasks such as text summarization \cite{shi_2021} and machine translation \cite{vaswani_2017}.

While these models have shown impressive results, they often require a large amount of training data and computing resources. However, only a few African languages have benefited from these advancements due to inadequate data. To address this issue, researchers have proposed transfer learning-based approaches, where a pretrained model is finetuned for a specific NLG task. 
Transfer learning \cite{JMLR:v21:20-074, he2022towards, ruder-etal-2019-transfer} has enabled the use of low-resource languages on various NLP tasks. Due to lack of adequate (or good quality) pretraining data \cite{caswell2021quality}, transfer learning is often the most accessible method for only a few low-resource languages leaving behind a vast majority of extremely low-resource languages. This is because about $90\%$ of the world's languages is claimed to be either \textit{left-behinds}, in that it is probably impossible to build NLP resources for them, or \textit{scraping-bys} with no labelled datasets \cite{joshi-etal-2020-state}. For the left-behinds, labelled and unlabelled data are unavailable and even transfer learning approaches are beyond reach while the scraping-by languages have no labelled data with which to evaluate model performance.

\subsection{Language Models}

Only a few African languages have benefited from the recent advancement of language models (LM) due to inadequate data sizes. We now describe encoder-decoder LMs that support NLP tasks in African languages. We describe these under two broad headings: massively multilingual models and African models. We summarize the models and African languages they cover in Table~\ref{tab:resources}. 

\noindent \textbf{Multilingual Models:} The massively multilingual models such as 
mBART~\citep{liu-etal-2020-multilingual-denoising}, MT0~\cite{muennighoff2022crosslingual}, and mT5~\citep{xue-etal-2021-mt5} are trained on several languages. However, in most cases, only a few African languages are represented. Among the mentioned models, mT0 is pretrained on the highest number of African languages ($n$=$13$).
 
\noindent \textbf{African Models.} \citet{adelani-etal-2022-thousand} use pretrained T5, mT5, and mBART models and develop AfriByT5, AfriMT5, AfriMBART respectively to investigate machine translation in zero-shot and out-of-domain settings. The authors experiment on $17$ African languages and demonstrate that further pretraining is effective for adding new languages to pretrained models.~\citet{jude-ogundepo-etal-2022-afriteva} train AfriTeVa, an encoder-decoder language model from scratch on $10$ African languages and English using similar training objectives like T5 model. 

\begin{table*}[h!]
\scriptsize
\centering
\resizebox{\textwidth}{!}{%
\begin{tabular}{clclc}
\toprule
\textbf{Category} &\textbf{LM} &\textbf{Lang/Total} &\textbf{African Languages} & \textbf{Families}  \\   \midrule
\multirow{3}{*}{\textbf{ Multilingual}} &MBART & $3$/$50$ & afr, swh, yor. & 2 \\
&MT\-0 & $14$/$101$ & afr, amh, hau, ibo, lin, mlg, nyj, orm, sot, & 4 \\
& & & sna, som, swh, xho, yor, and zul  \\
&MT\-5 & $12$/$101$ & afr, amh, nya, hau, ibo, mlg, sna,  som, swh, xho, yor, and zul & 3\\ \midrule

\multirow{5}{*}{\textbf{ African}}&AfriVeTa &$10$/$10$ & gaz, amh, Gahuza, hau, ibo, pcm, som, swa, tir, and yor. & 3 \\
&AfriMT5 & $17$/$17$ & bam, bbj, ewe, fon, hau, ibo, lug, luo, pcm, mos, swa, tsn, twi, wol, yor, zul. & 3 \\ 
&AfriByT5 & $17$/$17$ & bam, bbj, ewe, fon, hau, ibo, lug, luo, pcm, mos, swa, tsn, twi, wol, yor, zul. & 3\\ 
&AfriMBART & $17$/$17$ & afr, amh, nya, hau, orm, som, swh, xho. & 3\\ \cmidrule{2-5}


& \textbf{\ourmodelnlg}\includegraphics[scale=0.04]{images/cheetah_paper_logo.png} &$517$/$517$ & Includes $517$ African languages. & $14$\\
\bottomrule
\end{tabular}
}
\caption{Comparing with available encoder-decoder models with African languages represented. \textbf{Lang/Total}. describe the number of African languages comparing with the covered languages in the pretrained language models. \textbf{Families}. describes the number of covered language families.  }
\label{tab:resources}

\end{table*}

\noindent \textbf{African Natural Language Understanding.} Several works attempt to improve the performance on African NLU tasks by proposing multilingual and African-dedicated models such as mBERT~\cite{devlin-etal-2019-bert}, XLM-R~\cite{conneau-etal-2020-unsupervised}, AfriBERTa~\cite{ogueji-etal-2021-small}, AfroLM~\cite{afroLM}, Afro-XLM-R~\cite{alabi-etal-2022-adapting}, KINYaBERT~\cite{nzeyimana-niyongabo-rubungo-2022-kinyabert}, and SERENGETI~\cite{adebara2023serengeti}.

\begin{table*}[!]
\scriptsize
\centering
\resizebox{\textwidth}{!}{%
\begin{tabular}{clp{3.5cm}cccc}
\toprule
\textbf{Category} & \textbf{Benchmark} & \textbf{Reference} & \textbf{Task} & \textbf{Lang/Total} & \textbf{Datasets} & \textbf{Tasks} \\ \toprule
\multirow{9}{*}{\rotatebox[origin=c]{90}{\textbf{Multilingual}}} & FLoRES200 & \cite{nllb2022} & 52/200 & MT & Wiki & 1 \\
& GEM\textsubscript{v1} & \cite{gehrmann-etal-2021-gem} & DRG, DT, RES, TS, SMP & 10/52 & 18 & 13 \\

& GEM\textsubscript{v2} & \cite{gehrmann-etal-2021-gem} & \begin{tabular}[c]{@{}l@{}}DRG, DT, PPH, QA,\\ RES, TS, SLG, SMP, TS\end{tabular} & 10/52 & 50 & 9\\

& IndicNLG & \cite{kumar-etal-2022-indicnlg} & BG, HG, SUM, PARA, QA & 0/11 & 5 & 5 \\
& IndoNLG & \cite{cahyawijaya-etal-2021-indonlg} & SUM, QA, Chit-Chat & 0/3 & 5 & 3 \\
& NLLB M.D. & \cite{nllb2022} & MT & 2/8 & Wiki & 1 \\
& NLLB S.D. & \cite{nllb2022} & MT & 2/8 & Wiki & 1 \\
& Toxicity200 & \cite{nllb2022} & MT & 50/200 & Wiki & 1 \\

& XGLUE & \cite{liang-etal-2020-xglue} & \begin{tabular}[c]{@{}l@{}}NER, POS, MLQA, PAWS-X,\\XLNI, NC, QADSM, WPR, \\QAM, QG, NTG \end{tabular} & 1/19 & 19 & 11\\ \midrule
\multirow{3}{*}{\rotatebox[origin=c]{90}{\textbf{African}}} & AfroMT & \cite{reid21afromt} & MT & 8/8 & 5 & 1 \\
& Menyo-20k & \cite{adelani-etal-2021-effect} & MT & 1/2 & 6 & 1 \\ \cmidrule{2-7}
& AfroNLG & Our Work & Cloze, CS, MT, QA, TG, SUM, PARA & 517/527 & 67 & 7 \\ \bottomrule
\end{tabular}%
}
\caption{A Comparison of AfroNLG with other multilingual Benchmarks. \textbf{MT}: Machine translation, \textbf{QA:} Question Answering, \textbf{CS:} Code-Switching, \textbf{TG}: Title Generation, \textbf{SUM}: Summarization, \textbf{PARA}: Paraphrase, \textbf{NER}: Named Entity Recognition, \textbf{POS}: Part-Of-Speech Tagging, \textbf{MLQA:} Multilingual Question Answering, \textbf{PAWS-X}: Parallel Aggregated Word Scrambling for Cross-Lingual Understanding, \textbf{XNLI}: Cross-Lingual Natural Language Interference, \textbf{NC}: News Classification, \textbf{QADSM}: Query-AD Matching, \textbf{WPR:} Web Page Ranking, \textbf{QAM}: QA Matching, \textbf{NTG}: News Title Generation, \textbf{BG}: WikiBio Biography Generation, and \textbf{HG:} Headline Generation. \textbf{SD}: Seed Data, \textbf{MD}: Multi Domain. \textbf{DRG:} Dialogue Response Generator, \textbf{DT:} Data-to-Text, \textbf{RES:} Reasoning, \textbf{TS:} Text Summarization, \textbf{SMP:} Text Simplification, \textbf{PPH:} Paraphrase, \textbf{SLG:} Slide Generation}
\label{tab:benchmarkcompare}
\end{table*}

\subsection{Benchmarks}
Multiple benchmarks have been developed for NLG. However, only a few of Africa's $2,000$ languages have been supported to date. In most cases, the benchmarks support only the machine translation task. We provide a brief overview under two headings: African and multilingual. We summarize key information about each benchmark in Table \ref{tab:benchmarkcompare}.

\noindent \textbf{African Benchmarks.} \texttt{AfroMT}~\cite{reid21afromt} is a multilingual machine translation benchmark. It consists of translation tasks between English and eight African languages — Afrikaans, Xhosa, Zulu, Rundi, Sesotho, Swahili, Bemba, and Lingala. \texttt{Menyo-20k} \cite{adelani-etal-2021-effect} is an MT evaluation benchmark for English-Yor\`{u}b\'{a}. 

\noindent \textbf{Multilingual with African Languages.} \texttt{FLoRES-200} \cite{nllb2022, guzman-etal-2019-flores} is an evaluation benchmark that provides MT evaluation support in $200$ languages including $52$ African languages. \texttt{GEM} \cite{gehrmann-etal-2021-gem, gehrmann2022gemv2} referenced as ``living" benchmark, comprises of $40$ tasks and supports $52$ languages including $10$ African languages.  \texttt{NLLB Seed Data}~\cite{nllb2022} is a set of professionally-translated sentences sampled from Wikipedia. It consists of around six thousand sentences in $39$ languages which include $8$ African language. 
Similarly, \texttt{NLLB Multi Domain}~\cite{nllb2022} is an MT evaluation benchmark made from a set of professionally-translated sentences in the news and health domains. It consists of approximately $3,000$ sentences in each domain and supports $8$ languages including $2$ African languages. \texttt{Toxicity-200}~\cite{nllb2022} is an evaluation benchmark to evaluate the presence of toxic items in the MT text. It provides support for $50$ African languages. \texttt{XGLUE}~\cite{liang-etal-2020-xglue} is a cross-lingual, multi-task benchmark created with multilingual and bilingual corpora. It supports $19$ languages and one African language, i.e., Swahili. 





\section{\ourmodelnlg}\label{sec:afrot5}
\subsection{Pretraining Data}
We are guided by three main principles in developing this data: quality, linguistic diversity, and coverage. 

\noindent \textbf{Quality.} Developing NLP technologies for low resource languages poses a significant challenge due to the limited availability of high-quality training data. To address this issue, we undertook the task of manually curating a diverse corpus spanning multiple domains, including news articles, health documents, religious texts, legal documents, and social media feeds. This manual curation approach was necessary because there were no existing datasets available for the majority of the languages we aimed to support, and we wanted to ensure the utilization of reliable and high-quality data. 

\noindent \textbf{Coverage.} In all, we train \ourmodelnlg~using a 42G multi-domain corpus across $517$ African languages and language varieties. The languages are spoken in $50$ of $54$ African countries and they are written with five scripts. This provides support to at least $500$M Africans.

\noindent \textbf{Linguistic Diversity.} The inclusion of languages from various domains, geographical regions, and linguistic typologies, along with the utilization of reliable data sources, contributes to enhancing the robustness and quality of \ourmodelnlg. Our data consists of languages from $14$ language families in Africa written in five different orthographies. Furthermore, our data spans languages with a vast array of exotic linguistic features including tone, vowel and consonant harmony, reduplication, word orders, and word classes.  

We provide further details on the data used for pretraining in Section \ref{app:pretraing-data} in the Appendix.


\subsection{Implementation Details}
\noindent\textbf{Vocabulary.}
We use SentencePiece \cite{kudo-richardson-2018-sentencepiece} to encode text as WordPiece tokens \cite{sennrich-etal-2016-neural} with $250$K WordPieces.  We also include data covering the ten top spoken languages globally: Arabic, English, French, German, Greek, Italian, Portuguese, Russian, Spanish, and Turkish. We use Wikipedia dumps for these ten languages. We use $1$M sentences for each language. However, we only include it in the vocabulary. 

\noindent \textbf{Models Architecture.}
We pretrain \ourmodelnlg~using the encoder-decoder architecture \cite{xue-etal-2021-mt5}. Each of the encoder and decoder components is similar in size and configuration to T5, with $12$ layers each with $12$ attention heads, and $768$ hidden units for the base model. In total, this results in a model with $\sim580$ million parameters. We provide further details in Table \ref{tab:modelsinfo}.
\begin{table*}[h!]
\scriptsize
\centering
\resizebox{\textwidth}{!}{%
\begin{tabular}{lccccccccccc} 
\toprule
\textbf{Model}            & \textbf{Size} & \textbf{Params} & \textbf{No.\_heads} & \textbf{No.\_layers} & \textbf{D\_model} & \textbf{Vocab}              & \textbf{S.\_Len} & \textbf{B. Size}   & \textbf{\#Train\_Steps}& \textbf{\#Langs}& \textbf{\#A.Langs} \\ \toprule

mT0      & base          & 580M            & 12                  & 12                   & 768            & $\sim$250k & 1024    & 1024 & UNK  & 101&  13\\
mT5      &  base          & 580M            & 12                  & 12                   & 768        & 250K       & 1024    & 1024 & 1M & 101    & 13\\            
AfriMT5 &base          & 580M            & UNK                  & UNK                   & UNK          & UNK       & UNK    & 2048                   & UNK  & 17 & 17\\
AfriTeVa &base          & 229M            & 12                  & 12                   & 768          & 40K       & 512    & 256                   & 500K  & 10 & 10\\
                      \midrule  
\textbf{\ourmodelnlg}\includegraphics[scale=0.04]{images/cheetah_paper_logo.png}  & base          & 580M            & 12                  & 12                   & 768               & 250K      & 1024    & 1024 & 1M  & 527 &  517                    \\

                          \bottomrule                   \end{tabular}
}
\caption{Parameters of \ourmodelnlg~ compared with other models.}
\label{tab:modelsinfo}
\end{table*}

\noindent{\textbf{Objective.}} We use an unsupervised (denoising) objective. The main idea is to feed the model with masked (corrupted) versions of the original sentence, and train it to reconstruct the original sequence. The denoising objective \cite{xue-etal-2021-mt5} works by randomly sampling and dropping out 15\% of tokens in the input sequence. All consecutive spans of dropped-out tokens are then replaced by a single sentinel token.

\noindent \textbf{Pretraining Procedure}
For pretraining \ourmodelnlg, we use a learning rate of $0.01$, a batch size of $1,024$ sequences, and a maximum sequence length of $1,024$. We pretrain each model for $1$M steps. We train our models on Google Cloud TPU with $128$ cores (v$3-128$) from TensorFlow Research Cloud (TFRC).\footnote{\href{https://sites.research.google/trc/about/}{https://sites.research.google/trc/about/}}

\section{AfroNLG Benchmark}\label{sec:benchmark}

We create AfroNLG, a multi-lingual, multi-task benchmark comprising $67$ test sets across six task clusters. Specifically, AfroNLG includes the following: cloze tasks, machine translation, paraphrase, question answering, summarization, and title generation. AfroNLG supports $527$ languages, including $517$ African languages and language varieties and the top $10$ world languages. To the best of our knowledge, this is the most extensive benchmark till date for African languages. Table \ref{tab:benchmarkcompare} shows, at a glance, how our benchmark compares to others benchmark. We provide the details of each task cluster and datasets in what follows. For detailed statistics about the task clusters, we refer to Appendix~\ref{app_sec:benchmark}.

\begin{figure}[t]
  \centering
  \includegraphics[width=0.95\columnwidth]{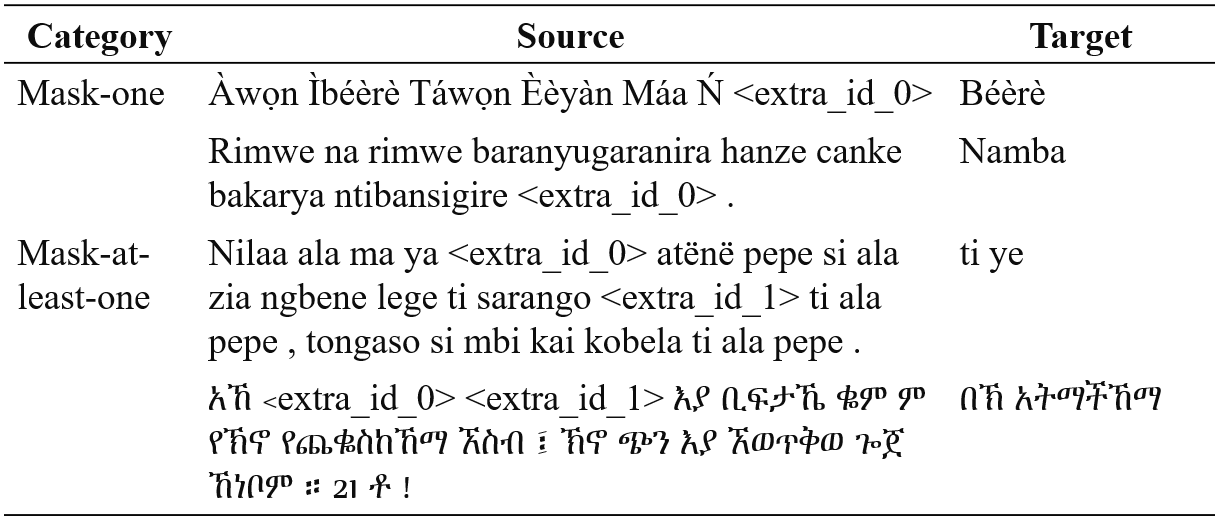}
\caption{Examples from the mask-one and mask-at-least-one cloze task data.} 
\label{fig:cloze_example} 
\end{figure}

\noindent \textbf{Cloze Test.} In order to comprehensively evaluate \ourmodelnlg~across all the languages it was pretrained on, we employ cloze-tasks as our evaluation approach and perform two cloze tasks experiments. These tasks assess the model's ability to fill in missing information. In the first cloze task, which we henceforth call \texttt{\textbf{mask-one}}, we randomly mask only one token in each sentence. In the second cloze-task, which we call \texttt{\textbf{mask-at-least-one}}, we randomly mask at least one token and not more than $10\%$ of the tokens in each sentence.  For each of the $517$ languages, we construct a cloze-task dataset comprising $200$ data points for each language in the Train set, $100$ examples for each language in the Test set, and $50$ data points for each language in the Dev set. We ensure that there is no overlap between the data used for the cloze tasks and the pretraining data. We show an example of our cloze task in Figure \ref{fig:cloze_example}.

\noindent \textbf{Machine Translation.} We include only datasets pertaining African languages in our benchmark. In selecting the languages for our MT benchmark, we strive to keep datasets that have been used in any published machine translation task. This allows us to cover a diverse set of languages and compare our models to existing SoTA across a large number of language pairs. Our benchmark thus contains data from Afro-MT\footnote{\href{https://github.com/machelreid/afromt}{https://github.com/machelreid/afromt}} \cite{reid-etal-2021-afromt}, Lafand-MT\footnote{\href{https://github.com/masakhane-io/lafand-mt}{https://github.com/masakhane-io/lafand-mt}} \cite{adelani-etal-2022-thousand}, PidginUNMT\footnote{\href{https://github.com/keleog/PidginUNMT}{https://github.com/keleog/PidginUNMT}} \cite{ogueji2019pidginunmt}, and SALT\footnote{\href{https://github.com/SunbirdAI/salt}{https://github.com/SunbirdAI/salt}} \cite{akera2022machine}. The datasets we consider make up $35$ language pairs. 


\noindent \textbf{Paraphrase.} A paraphrase task aims to create semantically similar and fluent paraphrases given an input text \cite{10.1162/tacl_a_00542, PALIVELA2021100025}. We use the TaPaCo dataset \cite{scherrer_yves_2020_3707949} for our paraphrase generation benchmark. TaPaCo is a freely available paraphrase corpus for $73$ languages extracted from the Tatoeba database. The dataset has four African languages: Afrikaans, Berber (a macro-language), Amazigh, and Kirundi. 



\noindent \textbf{Question Answering.} The QA task aims to provide answers to questions based on a knowledge base also referred to as contexts. We use TYDIA\footnote{\href{https://github.com/google-research-datasets/tydiqa}{https://github.com/google-research-datasets/tydiqa}} QA dataset \cite{clark-etal-2020-tydi}. The dataset has a primary task and a gold passage task. In our benchmark, we only include the gold passage task, where a correct answer is predicted from a passage containing one answer, similar to the existing reading comprehension task. 

\noindent \textbf{Summarization.} Summarization is the task of generating an abridged version of a text, while capturing the salient ideas and the intended information from the original text~\cite{nallapati-etal-2016-abstractive, king-etal-2022-dont}. We use the subset of XL-Sum~\cite{hasan-etal-2021-xl}, an abstractive summarization dataset, that consists of African languages including Amharic, Hausa, Igbo, Kirundi, Oromo, Pidgin, Somali, Swahili, Tigrinya, and Yor\`{u}b\'{a}. We also develop new test sets using data we crawled from the web, which are non-overlapping with XL-Sum. Specifically, we crawl data from BBC and Voice of Africa (webpages) for Hausa, Ndebele, and Swahili. 

\noindent \textbf{Title Generation.} The title generation task returns a single sentence title for a given article. Similar to the summarization task, we use XL-SUM to create a news title generation dataset. We also collect a new test set for title generation across $15$ languages. The dataset comprises $\sim6,000$ BBC and Voice of Africa articles, non-overlapping with XL-Sum, and is particularly useful for zero-shot title generation. 

 \section{Evaluation and Results }\label{sec:evaluation}
\begin{table*}[!]
\scriptsize
\centering
\resizebox{\textwidth}{!}{%
\begin{tabular}{llllllll} \toprule
\textbf{Cluster}                            & \textbf{Task}                                              & \textbf{Metric} & \textbf{mT0}        &\textbf{ mT5}        & \textbf{Afri-MT5}    & \textbf{AfriTeVa}     & \textbf{Cheetah}        \\ \toprule
\multirow{39}{*}{\rotatebox[origin=]{90}{\textbf{Machine Translation (MT)}}}                & English $\rightarrow$ Afrikaans                    & Bleu   & \textbf{20.38}\textsuperscript{$\pm$0.3 }  & 12.35\textsuperscript{$\pm$1.1 }  & 7.12\textsuperscript{$\pm$2.67 }   & 7.75\textsuperscript{$\pm$1.67}    & 19.72\textsuperscript{$\pm$0.75}     \\
                                   & English $\rightarrow$ Bemba                        & Bleu   & 19.19\textsuperscript{$\pm$0.3 }  & 12.28\textsuperscript{$\pm$0.48 } & 11.73\textsuperscript{$\pm$12.3 }  & \textbf{20.5}\textsuperscript{$\pm$0.87 }    & 18.9\textsuperscript{$\pm$1.22 }     \\
                                   & English $\rightarrow$ Lingala                      & Bleu   & \textbf{15.98}\textsuperscript{$\pm$1.16 } & 14.12\textsuperscript{$\pm$0.56 } & 14.32\textsuperscript{$\pm$12.74 } & 13.88\textsuperscript{$\pm$1.04 }   & 9.64\textsuperscript{$\pm$1.11}      \\
                                   & English $\rightarrow$ Rundi                        & Bleu   & \textbf{12.26}\textsuperscript{$\pm$0.47 } & 8.82\textsuperscript{$\pm$0.43 }  & 9.57\textsuperscript{$\pm$0.42 }   & 7.83\textsuperscript{$\pm$1.04 }    & 10.54\textsuperscript{$\pm$0.54 }     \\
                                   & English $\rightarrow$ Sesotho                      & Bleu   & 11.04\textsuperscript{$\pm$1.2 }  & 12.74\textsuperscript{$\pm$0.75 } & 10.0\textsuperscript{$\pm$1.79 }  & 10.76\textsuperscript{$\pm$1.4 }    & \textbf{13.3}\textsuperscript{$\pm$1.38 }      \\
                                   & English $\rightarrow$ Swahili                      & Bleu   & 10.59\textsuperscript{$\pm$1.84 } & 9.33\textsuperscript{$\pm$0.58 }  & 3.08\textsuperscript{$\pm$0.57 }   & 7.24\textsuperscript{$\pm$0.46 }    & \textbf{11.08}\textsuperscript{$\pm$0.61}     \\
                                   & English $\rightarrow$ Xhosa                        & Bleu   & 10.04\textsuperscript{$\pm$0.98 } & 8.25\textsuperscript{$\pm$0.7 }   & 3.86\textsuperscript{$\pm$1.35 }   & 7.5\textsuperscript{$\pm$0.32 }     & \textbf{12.34}\textsuperscript{$\pm$0.51 }     \\
                                   & English $\rightarrow$ Zulu                         & Bleu   & 17.65\textsuperscript{$\pm$1.86 } & 17.97\textsuperscript{$\pm$1.69 } & 1.9\textsuperscript{$\pm$1.11 }    & 13.45\textsuperscript{$\pm$1.81 }   & \textbf{19.49}\textsuperscript{$\pm$1.16 }     \\
                                   & English $\rightarrow$ Hausa                        & Bleu   & 5.06\textsuperscript{$\pm$0.21 }  & 4.96\textsuperscript{$\pm$0.16 }  & 0.85\textsuperscript{$\pm$0.04 }   & 7.32\textsuperscript{$\pm$0.00 }     & \textbf{9.22}\textsuperscript{$\pm$0.08 }     \\
                                   & English $\rightarrow$ Igbo                         & Bleu   & 13.05\textsuperscript{$\pm$0.17 } & 11.57\textsuperscript{$\pm$0.23 } & 1.12\textsuperscript{$\pm$0.09 }   & 12.34\textsuperscript{$\pm$0.23 }   & \textbf{16.75}\textsuperscript{$\pm$0.26 }     \\
                                   & English $\rightarrow$ Luganda                      & Bleu   & 2.17\textsuperscript{$\pm$2.77 }  & 3.33\textsuperscript{$\pm$0.35 }  & 0.09\textsuperscript{$\pm$0.01 }   & 4.21\textsuperscript{$\pm$0.77 }    & \textbf{9.75}\textsuperscript{$\pm$0.01 }      \\
                                   & English $\rightarrow$ N. Pidgin              & Bleu   & \textbf{33.17}\textsuperscript{$\pm$0.28 } & 32.65\textsuperscript{$\pm$0.19 } & 2.39\textsuperscript{$\pm$0.23 }   & 9.39\textsuperscript{$\pm$0.18}    & 32.64\textsuperscript{$\pm$0.14 }     \\
                                   & English $\rightarrow$ Swahili                      & Bleu   & 22.04\textsuperscript{$\pm$2.89 } & 23.2\textsuperscript{$\pm$0.23 }  & 2.79\textsuperscript{$\pm$0.08 }   & 22.39\textsuperscript{$\pm$0.28 }   & \textbf{28.11}\textsuperscript{$\pm$0.14 }     \\
                                   & English $\rightarrow$ Zulu                         & Bleu   & 6.83\textsuperscript{$\pm$0.29}  & 0.58\textsuperscript{$\pm$1.37 }  & 0.4\textsuperscript{$\pm$0.03 }    & 4.45\textsuperscript{$\pm$0.37 }    & \textbf{11.75}\textsuperscript{$\pm$0.38 }     \\
                                   & English $\rightarrow$ Twi                          & Bleu   & 3.4\textsuperscript{$\pm$0.12 }   & 1.23\textsuperscript{$\pm$0.03}  & 0.03\textsuperscript{$\pm$0.0}   & 1.68\textsuperscript{$\pm$0.94 }    & \textbf{4.64}\textsuperscript{$\pm$0.13 }      \\
                                   & English $\rightarrow$ Yoruba                       & Bleu   & 5.42\textsuperscript{$\pm$0.85 }  & 2.58\textsuperscript{$\pm$3.1 }   & 0.04\textsuperscript{$\pm$0.0}    & 3.63\textsuperscript{$\pm$4.01 }    & \textbf{7.83}\textsuperscript{$\pm$0.14 }      \\
                                   & English $\rightarrow$ Zulu                         & Bleu   & 10.28\textsuperscript{$\pm$0.49 } & 1.31\textsuperscript{$\pm$2.26 }  & 0.14\textsuperscript{$\pm$0.03 }   & 3.8\textsuperscript{$\pm$4.2 }     & \textbf{12.13}\textsuperscript{$\pm$0.1 }      \\
                                   & French $\rightarrow$ Bambara                       & Bleu   & 2.0\textsuperscript{$\pm$2.6 }    & 0.37\textsuperscript{$\pm$0.19 }  & 0.15\textsuperscript{$\pm$0.01 }   & \textbf{3.18}\textsuperscript{$\pm$0.18 }    & 3.06\textsuperscript{$\pm$0.27 }      \\
                                   & French $\rightarrow$ Ghomálá’                      & Bleu   & 0.4\textsuperscript{$\pm$0.09 }   & 0.33\textsuperscript{$\pm$0.01 }  & 0.07\textsuperscript{$\pm$0.0}    & \textbf{0.96}\textsuperscript{$\pm$0.01 }    & 0.28\textsuperscript{$\pm$0.25 }     \\
                                   & French $\rightarrow$ Ewe                           & Bleu   & 0.7\textsuperscript{$\pm$0.35 }   & 0.31\textsuperscript{$\pm$0.36 }  & 0.09\textsuperscript{$\pm$0.07 }   & 0.84\textsuperscript{$\pm$0.16 }    & \textbf{3.47}\textsuperscript{$\pm$0.03 }      \\
                                   & French $\rightarrow$ Fon                           & Bleu   & 0.69\textsuperscript{$\pm$0.31 }  & 0.8\textsuperscript{$\pm$0.13 }  & 1.52\textsuperscript{$\pm$0.06 }   & \textbf{1.73}\textsuperscript{$\pm$0.53}    & 1.29\textsuperscript{$\pm$0.16 }      \\
                                   & French $\rightarrow$ Moore                         & Bleu   & 0.27\textsuperscript{$\pm$0.06 }  & 0.12\textsuperscript{$\pm$0.05 }  & 0.19\textsuperscript{$\pm$0.02}   & 0.47\textsuperscript{$\pm$0.04 }    & \textbf{1.66}\textsuperscript{$\pm$0.86 }      \\
                                   & French $\rightarrow$ Wolof                         & Bleu   & 4.02\textsuperscript{$\pm$0.12 }  & 0.3\textsuperscript{$\pm$0.05 }   & 0.11\textsuperscript{$\pm$0.01}   & \textbf{3.08}\textsuperscript{$\pm$0.25 }    & 3.01\textsuperscript{$\pm$0.07 }     \\

                                & English $\rightarrow$ N. Pidgin (UNMT) & Bleu   & \textbf{27.44}\textsuperscript{$\pm$0.26 } & 23.42\textsuperscript{$\pm$1.61 } & 7.05\textsuperscript{$\pm$1.37 }   & 22.54\textsuperscript{$\pm$0.84 }   & 26.56\textsuperscript{$\pm$0.04 }      \\

                                   & Acholi $\rightarrow$ English                       & Bleu   & 16.41\textsuperscript{$\pm$0.08 } & 11.16\textsuperscript{$\pm$4.77 } & 4.9\textsuperscript{$\pm$0.11 }    & 8.37\textsuperscript{$\pm$8.12 }    & \textbf{19.33}\textsuperscript{$\pm$0.1 }      \\
                                   & Acholi $\rightarrow$ Lugbara                       & Bleu   & 2.57\textsuperscript{$\pm$0.21 }  & 1.48\textsuperscript{$\pm$1.31 }  & 2.44\textsuperscript{$\pm$0.37 }   & \textbf{8.29}\textsuperscript{$\pm$0.14 }    & 7.21\textsuperscript{$\pm$0.69 }      \\
                                   & Acholi $\rightarrow$ Luganda                       & Bleu   & 3.64\textsuperscript{$\pm$0.07 }  & 1.74\textsuperscript{$\pm$0.12 }  & 0.92\textsuperscript{$\pm$0.01 }   & 5.53\textsuperscript{$\pm$0.34 }    & \textbf{8.03}\textsuperscript{$\pm$0.38 }      \\
                                   & Acholi $\rightarrow$ Nyankore                      & Bleu   & 2.17\textsuperscript{$\pm$0.14 }  & 0.79\textsuperscript{$\pm$0.51 }  & 0.46\textsuperscript{$\pm$0.03 }   & 4.26\textsuperscript{$\pm$0.54 }    & \textbf{5.1}\textsuperscript{$\pm$0.14 }  \\
                                   & Acholi $\rightarrow$ Ateso                         & Bleu   & 1.64\textsuperscript{$\pm$2.34 }  & 1.94\textsuperscript{$\pm$0.25}  & 4.9\textsuperscript{$\pm$0.11 }    & \textbf{7.74}\textsuperscript{$\pm$0.33 }    & 6.33\textsuperscript{$\pm$0.6 }       \\
                                   & English $\rightarrow$ Lugbara                      & Bleu   & 6.19\textsuperscript{$\pm$6.33 }  & 8.38\textsuperscript{$\pm$0.49 }  & 5.93\textsuperscript{$\pm$0.22 }   & 10.95\textsuperscript{$\pm$0.32 }   & \textbf{11.61}\textsuperscript{$\pm$0.28 }     \\
                                   & English $\rightarrow$ Luganda                      & Bleu   & 12.08\textsuperscript{$\pm$0.03 } & 10.58\textsuperscript{$\pm$0.25 } & 2.59\textsuperscript{$\pm$0.73 }   & 12.41\textsuperscript{$\pm$0.35 }   & \textbf{17.12}\textsuperscript{$\pm$0.16 }    \\
                                   & English $\rightarrow$ Nyankore                     & Bleu   & 6.46\textsuperscript{$\pm$0.08 }  & 5.69\textsuperscript{$\pm$0.02 }  & 1.4\textsuperscript{$\pm$0.39 }    & 7.88\textsuperscript{$\pm$0.18 }    & \textbf{9.04}\textsuperscript{$\pm$0.24 }      \\
                                   & English $\rightarrow$ Ateso (salt)                 & Bleu   & 10.24\textsuperscript{$\pm$0.06 } & 8.28\textsuperscript{$\pm$0.19 }  & 4.91\textsuperscript{$\pm$0.59 }   & \textbf{11.64}\textsuperscript{$\pm$0.49 }   & 11.12\textsuperscript{$\pm$0.38 }     \\
                                   & Lugbara $\rightarrow$ Ateso                        & Bleu   & 2.21\textsuperscript{$\pm$0.35 }  & 1.5\textsuperscript{$\pm$0.2 }    & 2.22\textsuperscript{$\pm$0.15 }   & \textbf{6.67}\textsuperscript{$\pm$0.32 }    & 3.68\textsuperscript{$\pm$0.31 }      \\
                                   & Luganda $\rightarrow$ Lugbara                      & Bleu   & 3.96\textsuperscript{$\pm$0.57 }  & 2.61\textsuperscript{$\pm$0.12 }  & 3.44\textsuperscript{$\pm$0.32 }   & \textbf{8.05}\textsuperscript{$\pm$0.23 }    & 7.99\textsuperscript{$\pm$0.47 }      \\
                                   & Luganda $\rightarrow$ Ateso                        & Bleu   & 4.47\textsuperscript{$\pm$0.08 }  & 3.01\textsuperscript{$\pm$0.16 }  & 2.5\textsuperscript{$\pm$0.22 }    & \textbf{8.17}\textsuperscript{$\pm$0.18 }    & 8.13\textsuperscript{$\pm$0.33 }      \\
                                   & Nyankore $\rightarrow$ Lugbara                     & Bleu   & 3.45\textsuperscript{$\pm$0.29 }  & 2.1\textsuperscript{$\pm$0.32 }   & 2.6\textsuperscript{$\pm$0.29 }    & \textbf{7.5}\textsuperscript{$\pm$0.09 }     & 7.29\textsuperscript{$\pm$0.09 }     \\
                                   & Nyankore $\rightarrow$ Luganda                     & Bleu   & 8.54\textsuperscript{$\pm$0.17 }  & 6.91\textsuperscript{$\pm$0.23 }  & 2.01\textsuperscript{$\pm$0.25 }   & \textbf{6.77}\textsuperscript{$\pm$6.73 }    & 6.25\textsuperscript{$\pm$10.26 }     \\
                                   & Nyankore $\rightarrow$ Ateso                       & Bleu   & 3.33\textsuperscript{$\pm$0.11 }  & 2.25\textsuperscript{$\pm$0.23 }  & 2.12\textsuperscript{$\pm$0.4 }    & 6.27\textsuperscript{$\pm$0.12 }    & \textbf{6.36}\textsuperscript{$\pm$0.4 }      \\ \midrule
\multirow{3}{*}{\textbf{Paraphrase}}       &  Multilingual                           & Bleu   & 41.79\textsuperscript{$\pm$0.28 } & 41.75\textsuperscript{$\pm$0.21 } & 34.72\textsuperscript{$\pm$0.51 }  & 43.02\textsuperscript{$\pm$1.25 }  & \textbf{43.23}\textsuperscript{$\pm$0.09 }     \\
                                   &  Berber                                 & Bleu   & 44.84\textsuperscript{$\pm$0.31 } & 44.03\textsuperscript{$\pm$0.24 } & 36.08\textsuperscript{$\pm$0.83}  & \textbf{**46.4}1\textsuperscript{$\pm$0.71 } & 46.0\textsuperscript{$\pm$0.27 }      \\
                                   &  Kabyle                                 & Bleu   & 25.91\textsuperscript{$\pm$0.13 } & 25.32\textsuperscript{$\pm$0.46 } & 11.56\textsuperscript{$\pm$0.73}  & 16.06\textsuperscript{$\pm$14.79 }  & \textbf{26.27}\textsuperscript{$\pm$0.56 }     \\ \midrule
{\textbf{Question Answering}}                 & QA Swahili                                       & F1     & \textbf{79.84}\textsuperscript{$\pm$0.19 } & 72.04\textsuperscript{$\pm$0.54 } & 0           & 62.64\textsuperscript{$\pm$0.78 }   & 71.98\textsuperscript{$\pm$1.18 }    \\ \midrule
\multirow{11}{*}{\rotatebox[origin=]{90}{\textbf{Summarization}}}    &  Multilingual                            & RougeL & 22.31\textsuperscript{$\pm$0.12 } & 22.23\textsuperscript{$\pm$0.04 } & 5.34\textsuperscript{$\pm$0.48 }     & 18.97\textsuperscript{$\pm$0.06 }   & \textbf{24.86}\textsuperscript{$\pm$0.02 }     \\
                                   &  Amharic                                 & RougeL & 13.81\textsuperscript{$\pm$0.04 } & 13.09\textsuperscript{$\pm$0.03 } & 4.4\textsuperscript{$\pm$1.07}    & 8.29\textsuperscript{$\pm$0.51 }    & \textbf{15.09}\textsuperscript{$\pm$0.1 }      \\
                                   &  Igbo                                    & RougeL & \textbf{18.9}\textsuperscript{$\pm$0.73 }  & 13.22\textsuperscript{$\pm$0.46 } & 14.24\textsuperscript{$\pm$0.39 }  & 16.05\textsuperscript{$\pm$0.49 }   & 17.36\textsuperscript{$\pm$0.43 }     \\
                                   &  Oromo                                   & RougeL & 11.28\textsuperscript{$\pm$0.03 } & 10.51\textsuperscript{$\pm$0.07 } & 3.52\textsuperscript{$\pm$0.49}   & 7\textsuperscript{$\pm$1.73 }       & \textbf{14.53}\textsuperscript{$\pm$0.1 }      \\
                                   &  Rundi                                   & RougeL & 19.63\textsuperscript{$\pm$0.01 } & 18.02\textsuperscript{$\pm$0.13 } & 11.82\textsuperscript{$\pm$0.39 }  & 16.13\textsuperscript{$\pm$0.03 }  & \textbf{22.57}\textsuperscript{$\pm$0.04 }     \\
                                   &  Swahili                                 & RougeL & 26.38\textsuperscript{$\pm$0.02 } & 24.81\textsuperscript{$\pm$0.11 } & 15.07\textsuperscript{$\pm$0.17 }  & 21.59\textsuperscript{$\pm$0.13 }   & \textbf{29.05}\textsuperscript{$\pm$0.13 }     \\
                                   &  Yoruba                                  & RougeL & 21.57\textsuperscript{$\pm$0.05 } & 20.06\textsuperscript{$\pm$0.12 } & 13.52\textsuperscript{$\pm$0.18 }  & 17.3\textsuperscript{$\pm$0.11 }    & \textbf{22.49}\textsuperscript{$\pm$0.0 }     \\
                                   &  Hausa                                   & RougeL & 26.46\textsuperscript{$\pm$0.06 } & 25.76\textsuperscript{$\pm$0.02 } & 19.96\textsuperscript{$\pm$0.26 } & 25.19\textsuperscript{$\pm$0.11 }   & \textbf{30.07}\textsuperscript{$\pm$0.31 }     \\
                                   &  Nigerian Pidgin                         & RougeL & 26.54\textsuperscript{$\pm$0.05 } & 25.79\textsuperscript{$\pm$0.1 }  & 14.28\textsuperscript{$\pm$1.23 }  & 20.29\textsuperscript{$\pm$0.12 }   & \textbf{27.08}\textsuperscript{$\pm$0.02 }     \\
                                   &  Somali                                  & RougeL & 20.69\textsuperscript{$\pm$0.08 } & 19.21\textsuperscript{$\pm$0.06 } & 13.62\textsuperscript{$\pm$0.81 }  & 19.27\textsuperscript{$\pm$0.18 }   & \textbf{23.92}\textsuperscript{$\pm$0.04 }     \\
                                   &  Tigrinya                                & RougeL & 15.84\textsuperscript{$\pm$0.13 } & 13.93\textsuperscript{$\pm$0.11 } & 6.53\textsuperscript{$\pm$0.42 }   & 10.07\textsuperscript{$\pm$0.09 }   & \textbf{16.88}\textsuperscript{$\pm$0.12 }     \\ \midrule
\multirow{11}{*}{\rotatebox[origin=]{90}{\textbf{Title Generation}}}  & Multilingual                     & Bleu   & 6.53\textsuperscript{$\pm$0.02 }  & 6.65\textsuperscript{$\pm$0.08 }  & 0.1\textsuperscript{$\pm$0.02 }    & 5.2\textsuperscript{$\pm$0.02 }     & \textbf{7.52}\textsuperscript{$\pm$0.07 }      \\ 
                                   & Amharic                          & Bleu   & 3.13\textsuperscript{$\pm$0.23 }  & 2.65\textsuperscript{$\pm$0.68 }  & 0.34\textsuperscript{$\pm$0.14 }   & 2.31\textsuperscript{$\pm$0.14 }    & \textbf{4.34}\textsuperscript{$\pm$0.34 }      \\
                                   & Igbo                             & Bleu   & 6.95\textsuperscript{$\pm$0.13 }  & 6.9\textsuperscript{$\pm$0.22 }   & 0.77\textsuperscript{$\pm$0.12 }   & 4.61\textsuperscript{$\pm$0.14 }   & \textbf{8.47}\textsuperscript{$\pm$0.07 }      \\
                                   & Oromo                            & Bleu   & 1.1\textsuperscript{$\pm$1.84 }  & 2.66\textsuperscript{$\pm$0.19 } & 0.21\textsuperscript{$\pm$0.06 }  & 1.54\textsuperscript{$\pm$0.17 }    & \textbf{3.26}\textsuperscript{$\pm$0.21 }      \\
                                   & Rundi                            & Bleu   & 4.4\textsuperscript{$\pm$0.28 }   & 4.13\textsuperscript{$\pm$0.22 } & 0.84\textsuperscript{$\pm$0.07 }   & 3.33\textsuperscript{$\pm$0.23 }    & \textbf{6.05}\textsuperscript{$\pm$0.5 }       \\
                                   & Swahili                          & Bleu   & 9.1\textsuperscript{$\pm$0.23 }   & 9.31\textsuperscript{$\pm$0.11 }  & 1.22\textsuperscript{$\pm$0.09 }   & 7.01\textsuperscript{$\pm$0.09 }    & \textbf{10.59}\textsuperscript{$\pm$0.6 }     \\
                                   & Yoruba                           & Bleu   & 6.8\textsuperscript{$\pm$0.16 }   & 7.23\textsuperscript{$\pm$0.59 }  & 0.34\textsuperscript{$\pm$0.05}   & 5.04\textsuperscript{$\pm$2.0 }    & \textbf{7.97}\textsuperscript{$\pm$0.32 }      \\
                                   & Hausa                            & Bleu   & 8.11\textsuperscript{$\pm$0.24 }  & 7.3\textsuperscript{$\pm$0.34 }   & 2.59\textsuperscript{$\pm$0.01 }   & 6.69\textsuperscript{$\pm$0.18 }    & \textbf{8.48}\textsuperscript{$\pm$0.23}      \\
                                   & Nigerian Pidgin                  & Bleu   & \textbf{6.75}\textsuperscript{$\pm$0.6 }   & 3.96\textsuperscript{$\pm$4.3 }  & 0.89\textsuperscript{$\pm$0.02 }   & 4.72\textsuperscript{$\pm$0.84 }    & 6.22\textsuperscript{$\pm$0.28 }      \\
                                   & Somali                           & Bleu   & 3.37\textsuperscript{$\pm$0.21 } & 3.31\textsuperscript{$\pm$0.16 }  & 0.38\textsuperscript{$\pm$0.11 }   & 2.82\textsuperscript{$\pm$0.47 }    & \textbf{5.25}\textsuperscript{$\pm$0.14 }      \\
                                   & Tigrinya                         & Bleu   & 2.99\textsuperscript{$\pm$0.1 }   & 2.94\textsuperscript{$\pm$1.09 }  & 0.7\textsuperscript{$\pm$0.18 }    & 1.92\textsuperscript{$\pm$0.26 }    & \textbf{5.1}\textsuperscript{$\pm$0.05 }      \\   \midrule

                                   \multirow{3}{*}{\textbf{Cloze-task}}       
                                   & Mask-one & Bleu & 13.61\textsuperscript{$\pm$0.91 } & 8.18\textsuperscript{$\pm$3.94 } & 0.00\textsuperscript{$\pm$0.00 } & 8.36\textsuperscript{$\pm$3.42 }   & \textbf{13.98}\textsuperscript{$\pm$0.32 }    \\
                                   &  Mask-at-least-one                           & Bleu   & 2.36\textsuperscript{$\pm$0.11 } & 2.66\textsuperscript{$\pm$0.09 } & 0.93\textsuperscript{$\pm$0.12 } & 0.68\textsuperscript{$\pm$0.09 }   & \textbf{7.07}\textsuperscript{$\pm$0.09 }    \\
                                   \midrule
                                   & \multicolumn{2}{r}{\textbf{AfroNLG Score}}                          & 12.56     & 11.05     & 5.15       & 10.84        & \textbf{14.25} \\ \bottomrule
\end{tabular}%
}
    \caption{Average performance of finetuned African and multilingual models across three runs on AfroLNG benchmark test sets. }
    \label{tab:results}
\end{table*}
We evaluate \ourmodelnlg~on six task clusters of AfroNLG benchmark and compare to performance on mT0, mT5, Afri-MT5, and AfriTeVa. We report results in Table \ref{tab:results}. For all models, we finetune on the training data split (Train) for $20$ epochs with an early stopping of $5$ epochs, learning-rate of  $5e-5$, batch size of $16$, and sequence length of $512$. All experiments were performed on $4$ GPUs (Nvidia V100). We report the results of each experiment as an \textit{average of three runs}, each with a different seed.\footnote{Specifically, we use seed values $41$, $1512$, and $20235$.} For multilingual datasets in each task cluster, we show evaluation results per language. ~\ourmodelnlg~outperforms other models on many languages across the six task clusters. We provide detailed information of model performance next. 

\noindent{\textbf{Cloze Test.}}\label{subsec:cloze}
\ourmodelnlg~outperforms all other models on both cloze tasks as in Table \ref{tab:results}. We show the results for each language that is supported by the models compared in Table \ref{tab:cloze_mask_one} and Table \ref{tab:cloze_mask_at_least_one}. The performance of all models on mask-one is better than the performance on mask-at-least-one, reflecting how increasing the number of masked tokens makes the task more challenging. It is also important to mention that since evaluation is based on BLEU it does not reflect correct synonyms that each model may have generated to replace the masked tokens.

\noindent{\textbf{Machine Translation.}}
\ourmodelnlg~sets a new SOTA on $23$ tasks surpassing previous models. The mT0 and AfriTEVA models also demonstrate strong performance on six languages. Notably, pairs with French as the source language tend to yield the lowest BLEU scores, indicating relatively lower translation quality. On the other hand, the language pair involving English to Nigerian Pidgin, specifically on LafandMT and PidginUNMT, showcases the highest BLEU scores. We assume that the similarity between the Nigerian Pidgin and English contributes favourably to translation quality in these scenarios. We also report CHRF and CHRF++ results in Table \ref{tab:results_chrf} and Table \ref{tab:results_chrfplus} in the Appendix. 

\noindent{\textbf{Paraphrase}.}
In the three paraphrase tasks, \ourmodelnlg~demonstrates remarkable superiority over all other models. Specifically, we achieve an impressive ROUGE score of $46.0$ on the Berber paraphrase task, surpassing the second-best model by a margin of approximately two points.

\noindent{\textbf{Question Answering.}}
In the task of question answering, mT0 exhibits superior performance compared to both \ourmodelnlg~and other models. While mT5 achieves the second-highest performance, \ourmodelnlg~attains the third-highest performance in this task. 


\noindent{\textbf{Summarization.}}
\ourmodelnlg~sets a new SOTA on $11$ languages, outperforming other models by an average margin of at least three points. Detailed results can be found in Table \ref{tab:results}.

\noindent{\textbf{Title Generation.}}
On the Title generation task, \ourmodelnlg~sets a new SOTA on $11$ languages. We report results in Table \ref{tab:results}.

\subsection{Investigating linguistic capabilities}\label{sec:cfg}
In order to further test the utility of our models, we use grammar templates to construct test data in English. We use nine linguistic rules and $19$ lexical items to generate $152$ sentences. Next, we use our model to translate from source to target and manually evaluate the quality of the generated data. We design new evaluation metrics, \textit{faithfulness} and \textit{fluency}, for the manual evaluation. A detailed description follows.

\noindent{\textbf{Grammar templates.}} We use grammar templates \cite{mccoy-etal-2019-right} developed with context-free grammars (CFG) on the source side to construct controlled test sets in English. We use CFG on the source side alone because constituents and constituent order differs across languages.
We adopt this method for two reasons. First, utilizing grammar templates provides a standardized framework that ensures that the same grammatical phenomena are tested consistently. By employing a uniform approach, we can effectively isolate and evaluate specific linguistic features, facilitating a more rigorous and meaningful comparison of language model performance. Second, grammar templates exhibit a high degree of flexibility, allowing for easy modification and extension to encompass a wide range of linguistic phenomena. This adaptability not only facilitates the incorporation of new linguistic features but also enables the evolution of our test sets to match the dynamic landscape of natural language processing research.

Other alternatives to templates include using parsed corpora~\cite{bender-etal-2011-parser} or naturally occurring sentences. For the languages we explore, there are no good quality parsers, making automatic parsing inaccessible for this analysis. Furthermore, when a corpus is parsed automatically, the likelihood of encountering parsing errors escalates with the intricacy of the sentence structure~\cite{bender-etal-2011-parser, marvin-linzen-2018-targeted}. Conversely, if the test set exclusively comprises sentences with accurate gold parses, sourcing an ample quantity of instances showcasing syntactic complexities becomes an arduous task~\cite{marvin-linzen-2018-targeted}. Furthermore, the utilization of naturally occurring sentences introduces potential complications that might confound the interpretation of experiments~\cite{ettinger-etal-2018-assessing}.

The templates include transitive and intransitive structures, negative and affirmative structures, and structures with gender and number. Table \ref{tab:templates} provides examples of generated sentences using the templates\footnote{The entire generated grammar is available at our GitHub: \href{anonympus link}{anonymous link}. }. 

\begin{table}[!]
\scriptsize
\centering
\resizebox{0.7\columnwidth}{!}{%

\begin{tabular}{ll}
\toprule
\multicolumn{1}{l}{\textbf{Category}} & \multicolumn{1}{l}{\textbf{Example}} \\ \toprule
Intransitive                          & He left                              \\
Intransitive + Negation               & We did not leave                     \\
Transitive                            & You left Lagos                       \\
Transitive + Negation                 & She did not leave them        \\      \bottomrule
\end{tabular}%
}
   \caption{Some examples of sentences generated with the templates}
    \label{tab:templates}
\end{table}

\noindent{\textbf{Inference.}} We test three of our finetuned machine translation models with the generated dataset. This allows us to evaluate how much linguistic information the models have acquired during pretraining and finetuning. Specifically, we use the English$\rightarrow$Hausa, English$\rightarrow$Swahili, and English$\rightarrow$Yor\`{u}b\'{a} based on MT0, MT5, AfriTEVA, and \ourmodelnlg~models that were finetuned on the LafandMT dataset.  We do not include Afri-MT5 in this analysis because it has very low scores across several tasks as shown in Table \ref{tab:results}. Notably, Hausa, Swahili, and Yor\`{u}b\'{a} have distinct typologies and the performance of each model on each language gives further insight of performance across varying typological features (See Section \ref{app-sec:linguistic_details} for details). Table \ref{tab:grammar_distinction} shows some linguistic differences between the three languages. This method can be generalized to any African language. 

\begin{table}[!]
\scriptsize
\centering
\resizebox{\columnwidth}{!}{%
\begin{tabular}{lllll} 
\toprule
\multicolumn{1}{c}{\textbf{Lang.}} & \multicolumn{1}{l}{\textbf{Family}} & \multicolumn{1}{c}{\textbf{\# Tone}} & \textbf{Gender}       & \multicolumn{1}{c}{\textbf{Morphology}} \\ \toprule
Hausa                                 & Afro-Asiatic                        & Two                       & Two          & Isolating                               \\
Swahili &N.C. Bantu                  & None                          & Five & Agglutinative                           \\
Yourba & N.C. Non-Bantu              & Three                    & None         & Isolating                    \\ \bottomrule           
\end{tabular}%
}
   \caption{Some linguistic differences between Hausa, Swahili, and Yoruba. \textbf{N.C.} refers to Niger-Congo}
    \label{tab:grammar_distinction}
\end{table}


\subsection{Human evaluation}
To evaluate the effectiveness of each model across different languages, we assess the generated output's faithfulness and fluency using a five-point Likert scale. \textit{Faithfulness} measures how accurately a model's output corresponds to the input sentence, while \textit{fluency} assesses the grammatical coherence and plausibility of the generated output. We use both metrics because a model can produce coherent output that may not be faithful to the input sentence. This way, if faithfulness penalizes a model for outputs that are not true to the input or that include additional unnecessary information, fluency complements our evaluation of the quality of the same model if the output is fluent. For each grammar category, we return the average Likert point for each language and across the different models model.



\begin{figure*}
  \centering
  \begin{subfigure}[b]{0.49\textwidth}
    \includegraphics[width=\linewidth]{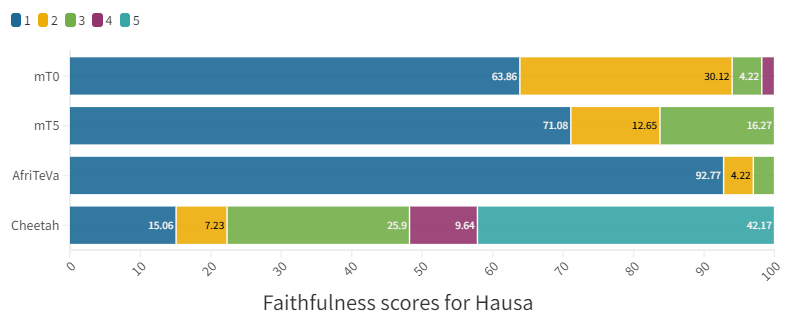}
  \end{subfigure}
  \hfill
   \begin{subfigure}[b]{0.49\textwidth}
    \includegraphics[width=\linewidth]{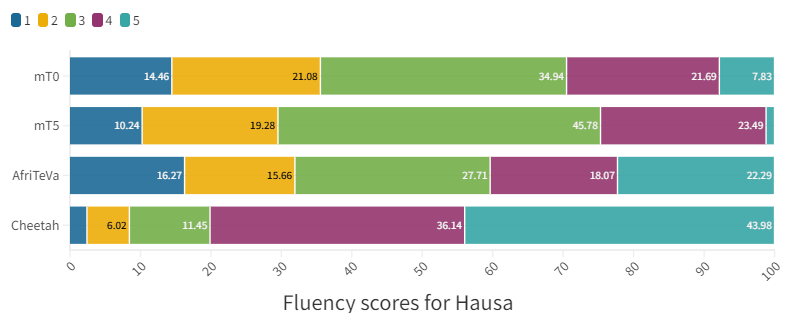}
  \end{subfigure}
  \hfill
   \begin{subfigure}[b]{0.49\textwidth}
    \includegraphics[width=\linewidth]{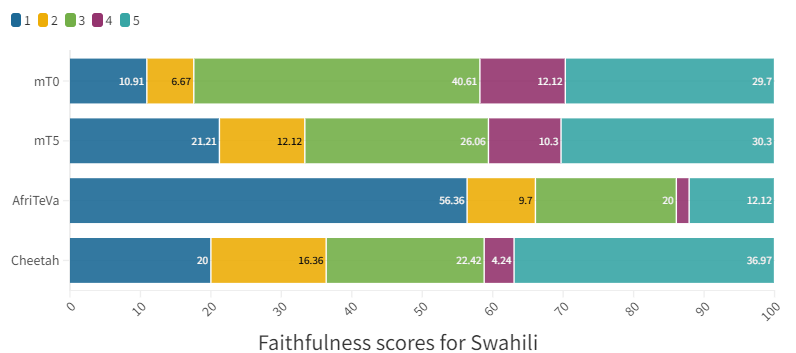}
 \end{subfigure}
 \hfill
  \begin{subfigure}[b]{0.49\textwidth}
    \includegraphics[width=\linewidth]{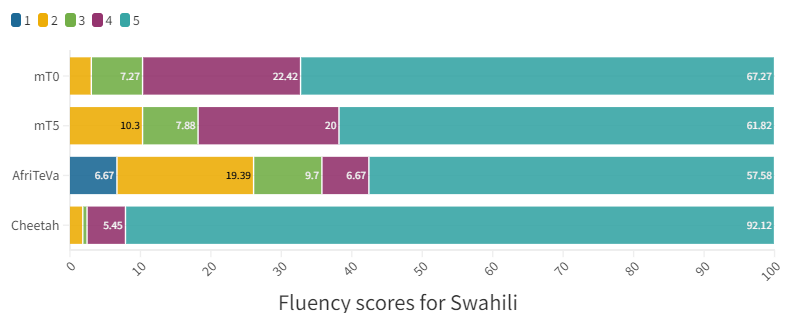}
  \end{subfigure}
 \hfill
  \begin{subfigure}[b]{0.49\textwidth}
    \includegraphics[width=\linewidth]{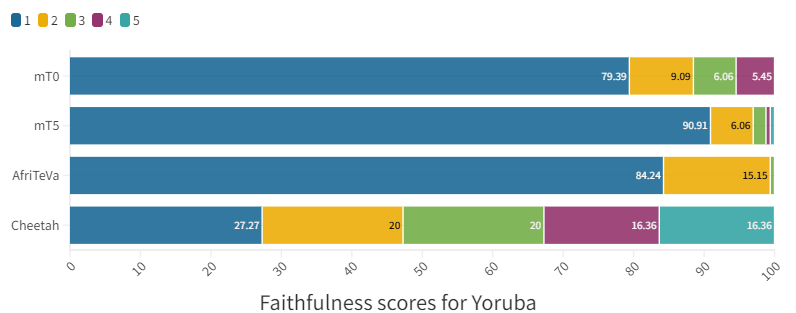}
 \end{subfigure} 
  \hfill
  \begin{subfigure}[b]{0.49\textwidth}
    \includegraphics[width=\linewidth]{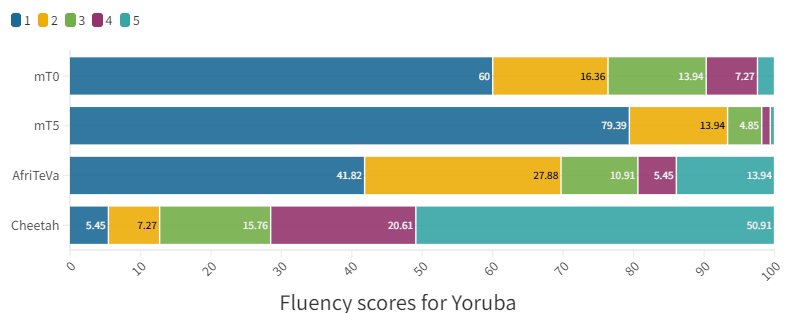}
  \end{subfigure}
  \caption{Faithfulness and fluency for Hausa, Swahili, and Yor\`{u}b\'{a}}\label{fig:faithfulness_fluency}
\end{figure*}

\subsection{Annotation}
We annotated each model's output for faithfulness and fluency. For Hausa and Yor\`{u}b\'{a}, two expert annotators evaluated the model's output for faithfulness and fluency. We ensured that each annotator has native speaker competency in reading and writing (while some had a linguistic background). We gave specific annotation instructions (See Section \ref{app-sec:annotation} in the Appendix) to ensure the values are not assigned arbitrarily. We also ensured that the annotators do not know who created which models to prevent any biases. We report the Kappa scores for inter-annotator agreement in Table~\ref{tab:interannottator}. For Swahili, only one annotator made it to the final annotation task since we could not acquire high quality annotations from other annotators. The Swahili annotator who did the final annotation is a university lecturer who teaches Swahili and has a Ph.D. in linguistics.

\begin{table}[]
\scriptsize
\centering
\resizebox{0.7\columnwidth}{!}{%
\begin{tabular}{lllll}\toprule
\multicolumn{1}{c}{\textbf{}} & \multicolumn{2}{c}{\textbf{hau}} & \multicolumn{2}{c}{\textbf{yor}} \\ \toprule
\textbf{Model}             & Faith.         & Flu.       & Faith.         & Flu.  \\ \midrule
mT0                         &   $90.54$        &    $97.62$           &                          $96.57$          &   $93.92$              \\
mT5                       &       $93.51$          &     $96.48$           &                    $82.23$      &   $81.10$               \\
AfriTeVa                       &    $87.27$              &   $96.94$            &       $88.56$            &  $84.73$             \\ 
Cheetah                       &      $96.61$        &    $97.26$           &    $87.11$            &    $92.64$             \\
\bottomrule   
\end{tabular}%
}
\caption{Kappa scores for Faithfulness (i.e., Faith.) and Fluency (i.e., Flu.) across the four models and three languages we evaluate. }\label{tab:interannottator}
\end{table}

\subsection{Fluency and Faithfulness Performance}
We report the distribution of faithfulness and fluency scores across all models and languages in Figure \ref{fig:faithfulness_fluency}. Overall, \ourmodelnlg~produces more faithful and more fluent outputs than other models on all languages. We now go on to provide detailed analysis of model performance. 

\noindent{\textbf{Intransitives}}
In the case of Hausa examples, all three models manage to produce intransitive examples. However, \ourmodelnlg~consistently appends objects to these intransitive examples. This inclination to add objects might stem from biases within the data used for pretraining or finetuning \ourmodelnlg. Nevertheless, it is worth noting that \ourmodelnlg~outperforms other models by generating more fluent and more faithful Hausa outputs. In the Swahili context, all models successfully generate intransitive translations, with model errors primarily related to tense. This performance discrepancy in Swahili can be attributed to its agglutinative structure, with models potentially lacking exposure to a comprehensive range of grammatical features during pretraining or finetuning. In the context of Yor\`{u}b\'{a}, all models consistently incorporate at least one object in each intransitive case. Notably, mT0 generates an output without an object approximately $5.88\%$ of the time. This may be because intransitive sentences inherently lack a clear direct object, making it more challenging for machine translation models to grasp context and select the accurate translation. In certain instances, some intransitive phrases can be polysemous, further complicating the translation process. Intransitive English verbs do not always retain their intransitive nature in Yor\`{u}b\'{a}. Furthermore, transitives with optional/truncated objects tend to have a  compulsory object in Yor\`{u}b\'{a}. This phenomenon potentially contributes to the models' tendency to append objects to intransitive Yor\`{u}b\'{a} phrases. For instance, whereas the intransitive "slept" in "John slept" maps to the intransitive form "John s\`{u}n" in Yor\`{u}b\'{a}, the intransitive verb "prayed", in "John prayed" becomes "John gbàdúrà", a transitive verb in Yor\`{u}b\'{a}. On the other hand, the transitive verb "ate" in "John ate", has an optional/truncated object in English but becomes "John j\textsubdot{e}un", a transitive with an obligatory object. In Yor\`{u}b\'{a}, both "ate" and "prayed" are transitive verbs that require an object. They are derived from "j\textsubdot{e}" (eat) and "o\'{u}nj\textsubdot{e}" (food), which give rise to "j\textsubdot{e}un" and "gb\`{a}" (collect) and "\`{a}d\'{u}r\`{a}" (prayer), resulting in "gb\`{a}d\'{u}r\`{a}" respectively.
We report the distribution of scores in Figure \ref{fig:intrans}.

\noindent{\textbf{Transitives}}
In the context of transitives, \ourmodelnlg~stands out as the top-performing model across all three languages, as illustrated in Figure \ref{fig:faithfulness_fluency}. \ourmodelnlg~demonstrates the capability to provide three distinct semantic senses for the polysemous transitive verb treated whereas the other models typically produce only a single semantic interpretation. In Swahili examples, certain instances exhibit the deletion or simplification of object markers in an ungrammatical manner. For a visual representation of the annotation of intransitive sentences in Yor\`{u}b\'{a}, please refer to Figure \ref{fig:intransitives}. Figure \ref{fig:trans} shows the distribution of model performance on transitives.


\noindent{\textbf{Negative}}
In the context of Yor\`{u}b\'{a}, all models are able to produce the correct negation marker including the correct tone marks. The tone patterns on negation markers may vary based on the context of words before and after the negation marker and it was interesting to see these variations in the models outputs. Despite this, mT0, MT5, and AfriTeVa have a tendency to output the negation of the antonym of the verb in each sentence rather than the negation of the verb. \ourmodelnlg~also makes similar mistakes about ~5\% of the time. 

\noindent{\textbf{Affirmative}}
The models generally perform better in the context of the affirmative examples than on the negated examples. However, in the context of Hausa, mT5, mT0, and AfriTeVa consistently output the antonym of the verb to be negated. For instance, the models return ``Sara left" rather than ``Sara did not leave". In the Swahili examples, we also find cases of double negation (which is not grammatically correct in Swahili). We show the distribution of results in Figure \ref{fig:trans_neg} and Figure \ref{fig:intrans_neg}.

\noindent{\textbf{Gender/Agreement}}
We find interesting cases of gender in the model's output. For example, whereas Yor\`{u}b\'{a} grammar does not distinguish gender, \ourmodelnlg~uses \textit{Ar\'{a}b\`{i}rin} (female) before every occurrence of the name ``Sara" to indicate that the it has a high probability of being feminine (see Figure \ref{fig:intransitives}). It is important to mention that ``Fred" is not annotated this way. For Hausa, which requires agreement between the gender of the noun and the verb, we find \ourmodelnlg~outperforming both mt0 and mt5 significantly. AfriTeVa, however, has very low accuracy in the context of gender. Furthermore, mt0, mt5, and \ourmodelnlg~return connotations for love and relationships for each examples where a male and female pronoun co-occur cross-lingually.

\noindent{\textbf{Number}}
\ourmodelnlg~significantly outperforms all three models in accurately assigning appropriate number markers. We also find that when translating the word "you" into Hausa, Swahili, or Yor\`{u}b\'{a}, all four models use either singular or plural forms. We assume that this is due to the fact that the second person in English (i.e., ``you") can be either singular or plural while each of these languages have a different word for the singular and plural forms. 

\section{Conclusion}\label{sec:conc}
In this work, we introduced~\ourmodelnlg, a massively multilingual language model designed for African natural language generation. We also propose a new African language generation benchmark, dubbed AfroNLG. Our evaluation benchmark is both sizeable and diverse. We evaluate \ourmodelnlg~on AfroNLG comparing it to three other models, two multilingual and one dedicated to African languages. The performance of \ourmodelnlg~surpasses that of all other models we evaluate. This is demonstrated by its superior AfroNLG score, which is approximately three times better than the combined performance of other models. Furthermore, \ourmodelnlg~outperforms all other models across $48$ out of $65$ test sets spanning six task clusters. We further analyze our model's robustness to lexical complexity and carry out human evaluation to inspect the model's perform on a controlled test set. Again, our results underscore superiority of our model.   
 \newpage
  \section{Limitations}\label{sec:limits}

We identify the following limitations for our work:
\begin{enumerate}
\item The limitations of our language model include the limited scope of our evaluation. Future work should focus on increasing the subset of languages evaluated manually in order to ensure quality. We believe automatic analyses are not sufficient for development of models that get deployed in particular applications.

 \item Another limitation is related to our inability to perform extensive analysis of biases and hateful speech present in our pretraining data. Again, this is due to relatively restricted access to native speakers (and even automated tools) to perform this analysis. As a result, we cannot fully ensure that our models are free from biases and socially undesirable effects. Therefore, it is important that these models be used with care and caution, and be analyzed for biases and socially undesirable effects before use.

\item Additionally, due to unavailability of sufficient computing resources, we were unable to evaluate larger multilingual language models. 

\end{enumerate}

 \section{Ethics Statement and Wider Impacts}\label{sec:ethics}
\ourmodelnlg~aligns with Afrocentric NLP where the needs of African people is put into consideration when developing technology. We believe \ourmodelnlg~will not only be useful to speakers of the languages supported, but also researchers of African languages such as anthropologists and linguists. We discuss below some use cases for~\ourmodelnlg~and offer a number of broad impacts. 
\begin{enumerate}
    \item \ourmodelnlg~aims to address the lack of access to technology in about $90\%$ of the world's languages, which automatically discriminates against native speakers of those languages. More precisely, it does so by focusing on Africa. To the best of our knowledge,~\ourmodelnlg~is the first massively multilingual PLM developed for African languages and language varieties. A model with knowledge of $517$ African languages, is by far the largest to date for African NLP. 
    \item \ourmodelnlg~enables improved access of important information to the African community in Indigenous African languages. This is especially beneficial for people who may not be fluent in other languages. This will potentially connect more people globally. 
    \item \ourmodelnlg~affords opportunities for language preservation for many African languages. To the best of our knowledge, \ourmodelnlg~consists of languages that have not been used for any NLP task until now. We believe that it can help encourage  continued use of these languages in several domains, as well as trigger future development of language technologies for many of these languages.
     \item Although LMs are useful for a wide range of applications, they can also be misused.~\ourmodelnlg~is developed using publicly available datasets that may carry biases. Although we strive to perform analyses and diagnostic case studies to probe performance of our models, our investigations are by no means comprehensive nor guarantee absence of bias in the data. In particular, we do not have access to native speakers of most of the languages covered. This hinders our ability to investigate samples from each (or at least the majority) of the languages.
 \end{enumerate}

\bibliography{custom}

\begin{thebibliography}{58}
\expandafter\ifx\csname natexlab\endcsname\relax\def\natexlab#1{#1}\fi

\bibitem[{Abadji et~al.(2021)Abadji, Su{\'a}rez, Romary, and Sagot}]{AbadjiOrtizSuarezRomaryetal.2021}
Julien Abadji, Pedro Javier~Ortiz Su{\'a}rez, Laurent Romary, and Beno{\^i}t Sagot. 2021.
\newblock \href {https://doi.org/10.14618/ids-pub-10468} {Ungoliant: An optimized pipeline for the generation of a very large-scale multilingual web corpus}.
\newblock Proceedings of the Workshop on Challenges in the Management of Large Corpora (CMLC-9) 2021. Limerick, 12 July 2021 (Online-Event), pages 1 -- 9, Mannheim. Leibniz-Institut f{\"u}r Deutsche Sprache.

\bibitem[{Adebara and Abdul-Mageed(2022)}]{adebara-abdul-mageed-2022-towards}
Ife Adebara and Muhammad Abdul-Mageed. 2022.
\newblock \href {https://doi.org/10.18653/v1/2022.acl-long.265} {Towards afrocentric {NLP} for {A}frican languages: Where we are and where we can go}.
\newblock In \emph{Proceedings of the 60th Annual Meeting of the Association for Computational Linguistics (Volume 1: Long Papers)}, pages 3814--3841, Dublin, Ireland. Association for Computational Linguistics.

\bibitem[{Adebara et~al.(2023)Adebara, Elmadany, Abdul-Mageed, and Inciarte}]{adebara2023serengeti}
Ife Adebara, AbdelRahim Elmadany, Muhammad Abdul-Mageed, and Alcides~Alcoba Inciarte. 2023.
\newblock \href {http://arxiv.org/abs/2212.10785} {Serengeti: Massively multilingual language models for africa}.

\bibitem[{Adelani et~al.(2022)Adelani, Alabi, Fan, Kreutzer, Shen, Reid, Ruiter, Klakow, Nabende, Chang, Gwadabe, Sackey, Dossou, Emezue, Leong, Beukman, Muhammad, Jarso, Yousuf, Niyongabo~Rubungo, Hacheme, Wairagala, Nasir, Ajibade, Ajayi, Gitau, Abbott, Ahmed, Ochieng, Aremu, Ogayo, Mukiibi, Ouoba~Kabore, Kalipe, Mbaye, Tapo, Memdjokam~Koagne, Munkoh-Buabeng, Wagner, Abdulmumin, Awokoya, Buzaaba, Sibanda, Bukula, and Manthalu}]{adelani-etal-2022-thousand}
David Adelani, Jesujoba Alabi, Angela Fan, Julia Kreutzer, Xiaoyu Shen, Machel Reid, Dana Ruiter, Dietrich Klakow, Peter Nabende, Ernie Chang, Tajuddeen Gwadabe, Freshia Sackey, Bonaventure F.~P. Dossou, Chris Emezue, Colin Leong, Michael Beukman, Shamsuddeen Muhammad, Guyo Jarso, Oreen Yousuf, Andre Niyongabo~Rubungo, Gilles Hacheme, Eric~Peter Wairagala, Muhammad~Umair Nasir, Benjamin Ajibade, Tunde Ajayi, Yvonne Gitau, Jade Abbott, Mohamed Ahmed, Millicent Ochieng, Anuoluwapo Aremu, Perez Ogayo, Jonathan Mukiibi, Fatoumata Ouoba~Kabore, Godson Kalipe, Derguene Mbaye, Allahsera~Auguste Tapo, Victoire Memdjokam~Koagne, Edwin Munkoh-Buabeng, Valencia Wagner, Idris Abdulmumin, Ayodele Awokoya, Happy Buzaaba, Blessing Sibanda, Andiswa Bukula, and Sam Manthalu. 2022.
\newblock \href {https://doi.org/10.18653/v1/2022.naacl-main.223} {A few thousand translations go a long way! leveraging pre-trained models for {A}frican news translation}.
\newblock In \emph{Proceedings of the 2022 Conference of the North American Chapter of the Association for Computational Linguistics: Human Language Technologies}, pages 3053--3070, Seattle, United States. Association for Computational Linguistics.

\bibitem[{Adelani et~al.(2021)Adelani, Ruiter, Alabi, Adebonojo, Ayeni, Adeyemi, Awokoya, and Espa{\~n}a-Bonet}]{adelani-etal-2021-effect}
David Adelani, Dana Ruiter, Jesujoba Alabi, Damilola Adebonojo, Adesina Ayeni, Mofe Adeyemi, Ayodele~Esther Awokoya, and Cristina Espa{\~n}a-Bonet. 2021.
\newblock \href {https://aclanthology.org/2021.mtsummit-research.6} {The effect of domain and diacritics in {Y}oruba{--}{E}nglish neural machine translation}.
\newblock In \emph{Proceedings of Machine Translation Summit XVIII: Research Track}, pages 61--75, Virtual. Association for Machine Translation in the Americas.

\bibitem[{Akera et~al.(2022)Akera, Mukiibi, Sanyu~Naggayi, Babirye, Owomugisha, Nsumba, Nakatumba-Nabende, Bainomugisha, Mwebaze, and Quinn}]{akera2022machine}
Benjamin Akera, Jonathan Mukiibi, Lydia Sanyu~Naggayi, Claire Babirye, Isaac Owomugisha, Solomon Nsumba, Joyce Nakatumba-Nabende, Engineer Bainomugisha, Ernest Mwebaze, and John Quinn. 2022.
\newblock Machine translation for african languages: Community creation of datasets and models in uganda.

\bibitem[{Alabi et~al.(2022)Alabi, Adelani, Mosbach, and Klakow}]{alabi-etal-2022-adapting}
Jesujoba~O. Alabi, David~Ifeoluwa Adelani, Marius Mosbach, and Dietrich Klakow. 2022.
\newblock \href {https://aclanthology.org/2022.coling-1.382} {Adapting pre-trained language models to {A}frican languages via multilingual adaptive fine-tuning}.
\newblock In \emph{Proceedings of the 29th International Conference on Computational Linguistics}, pages 4336--4349, Gyeongju, Republic of Korea. International Committee on Computational Linguistics.

\bibitem[{Becker(2002)}]{becker-2002-practical}
Tilman Becker. 2002.
\newblock \href {https://aclanthology.org/W02-2211} {Practical, template{--}based natural language generation with {TAG}}.
\newblock In \emph{Proceedings of the Sixth International Workshop on Tree Adjoining Grammar and Related Frameworks ({TAG}+6)}, pages 80--83, Universit{\'a} di Venezia. Association for Computational Linguistics.

\bibitem[{Bender et~al.(2011)Bender, Flickinger, Oepen, and Zhang}]{bender-etal-2011-parser}
Emily~M. Bender, Dan Flickinger, Stephan Oepen, and Yi~Zhang. 2011.
\newblock \href {https://aclanthology.org/D11-1037} {Parser evaluation over local and non-local deep dependencies in a large corpus}.
\newblock In \emph{Proceedings of the 2011 Conference on Empirical Methods in Natural Language Processing}, pages 397--408, Edinburgh, Scotland, UK. Association for Computational Linguistics.

\bibitem[{Brown et~al.(2020)Brown, Mann, Ryder, Subbiah, Kaplan, Dhariwal, Neelakantan, Shyam, Sastry, Askell, Agarwal, Herbert-Voss, Krueger, Henighan, Child, Ramesh, Ziegler, Wu, Winter, Hesse, Chen, Sigler, Litwin, Gray, Chess, Clark, Berner, McCandlish, Radford, Sutskever, and Amodei}]{brown_2020}
Tom~B. Brown, Benjamin Mann, Nick Ryder, Melanie Subbiah, Jared Kaplan, Prafulla Dhariwal, Arvind Neelakantan, Pranav Shyam, Girish Sastry, Amanda Askell, Sandhini Agarwal, Ariel Herbert-Voss, Gretchen Krueger, Tom Henighan, Rewon Child, Aditya Ramesh, Daniel~M. Ziegler, Jeffrey Wu, Clemens Winter, Christopher Hesse, Mark Chen, Eric Sigler, Mateusz Litwin, Scott Gray, Benjamin Chess, Jack Clark, Christopher Berner, Sam McCandlish, Alec Radford, Ilya Sutskever, and Dario Amodei. 2020.
\newblock Language models are few-shot learners.
\newblock In \emph{Proceedings of the 34th International Conference on Neural Information Processing Systems}, NIPS'20, Red Hook, NY, USA. Curran Associates Inc.

\bibitem[{Cahyawijaya et~al.(2021)Cahyawijaya, Winata, Wilie, Vincentio, Li, Kuncoro, Ruder, Lim, Bahar, Khodra, Purwarianti, and Fung}]{cahyawijaya-etal-2021-indonlg}
Samuel Cahyawijaya, Genta~Indra Winata, Bryan Wilie, Karissa Vincentio, Xiaohong Li, Adhiguna Kuncoro, Sebastian Ruder, Zhi~Yuan Lim, Syafri Bahar, Masayu Khodra, Ayu Purwarianti, and Pascale Fung. 2021.
\newblock \href {https://doi.org/10.18653/v1/2021.emnlp-main.699} {{I}ndo{NLG}: Benchmark and resources for evaluating {I}ndonesian natural language generation}.
\newblock In \emph{Proceedings of the 2021 Conference on Empirical Methods in Natural Language Processing}, pages 8875--8898, Online and Punta Cana, Dominican Republic. Association for Computational Linguistics.

\bibitem[{Chen et~al.(2023)Chen, Tam, Raffel, Bansal, and Yang}]{10.1162/tacl_a_00542}
Jiaao Chen, Derek Tam, Colin Raffel, Mohit Bansal, and Diyi Yang. 2023.
\newblock \href {https://doi.org/10.1162/tacl_a_00542} {{An Empirical Survey of Data Augmentation for Limited Data Learning in NLP}}.
\newblock \emph{Transactions of the Association for Computational Linguistics}, 11:191--211.

\bibitem[{Clark et~al.(2020)Clark, Choi, Collins, Garrette, Kwiatkowski, Nikolaev, and Palomaki}]{clark-etal-2020-tydi}
Jonathan~H. Clark, Eunsol Choi, Michael Collins, Dan Garrette, Tom Kwiatkowski, Vitaly Nikolaev, and Jennimaria Palomaki. 2020.
\newblock \href {https://doi.org/10.1162/tacl_a_00317} {{T}y{D}i {QA}: A benchmark for information-seeking question answering in typologically diverse languages}.
\newblock \emph{Transactions of the Association for Computational Linguistics}, 8:454--470.

\bibitem[{Conneau et~al.(2020)Conneau, Khandelwal, Goyal, Chaudhary, Wenzek, Guzm{\'a}n, Grave, Ott, Zettlemoyer, and Stoyanov}]{conneau-etal-2020-unsupervised}
Alexis Conneau, Kartikay Khandelwal, Naman Goyal, Vishrav Chaudhary, Guillaume Wenzek, Francisco Guzm{\'a}n, Edouard Grave, Myle Ott, Luke Zettlemoyer, and Veselin Stoyanov. 2020.
\newblock \href {https://doi.org/10.18653/v1/2020.acl-main.747} {Unsupervised cross-lingual representation learning at scale}.
\newblock In \emph{Proceedings of the 58th Annual Meeting of the Association for Computational Linguistics}, pages 8440--8451, Online. Association for Computational Linguistics.

\bibitem[{Costa-juss{\`a} et~al.(2022)Costa-juss{\`a}, Cross, {\c{C}}elebi, Elbayad, Heafield, Heffernan, Kalbassi, Lam, Licht, Maillard et~al.}]{nllb2022}
Marta~R Costa-juss{\`a}, James Cross, Onur {\c{C}}elebi, Maha Elbayad, Kenneth Heafield, Kevin Heffernan, Elahe Kalbassi, Janice Lam, Daniel Licht, Jean Maillard, et~al. 2022.
\newblock No language left behind: Scaling human-centered machine translation.
\newblock \emph{arXiv preprint arXiv:2207.04672}.

\bibitem[{Devlin et~al.(2019)Devlin, Chang, Lee, and Toutanova}]{devlin-etal-2019-bert}
Jacob Devlin, Ming-Wei Chang, Kenton Lee, and Kristina Toutanova. 2019.
\newblock \href {https://doi.org/10.18653/v1/N19-1423} {{BERT}: Pre-training of deep bidirectional transformers for language understanding}.
\newblock In \emph{Proceedings of the 2019 Conference of the North {A}merican Chapter of the Association for Computational Linguistics: Human Language Technologies, Volume 1 (Long and Short Papers)}, pages 4171--4186, Minneapolis, Minnesota. Association for Computational Linguistics.

\bibitem[{Dossou et~al.(2022)Dossou, Tonja, Yousuf, Osei, Oppong, Shode, Awoyomi, and Emezue}]{afroLM}
Bonaventure F.~P. Dossou, Atnafu~Lambebo Tonja, Oreen Yousuf, Salomey Osei, Abigail Oppong, Iyanuoluwa Shode, Oluwabusayo~Olufunke Awoyomi, and Chris~Chinenye Emezue. 2022.
\newblock \href {https://doi.org/10.48550/ARXIV.2211.03263} {Afrolm: A self-active learning-based multilingual pretrained language model for 23 african languages}.

\bibitem[{Dryer and Haspelmath(2013)}]{wals}
Matthew~S. Dryer and Martin Haspelmath, editors. 2013.
\newblock \href {https://wals.info/} {\emph{WALS Online}}.
\newblock Max Planck Institute for Evolutionary Anthropology, Leipzig.

\bibitem[{Du{\v{s}}ek and Jur{\v{c}}{\'\i}{\v{c}}ek(2015)}]{dusek-jurcicek-2015-training}
Ond{\v{r}}ej Du{\v{s}}ek and Filip Jur{\v{c}}{\'\i}{\v{c}}ek. 2015.
\newblock \href {https://doi.org/10.3115/v1/P15-1044} {Training a natural language generator from unaligned data}.
\newblock In \emph{Proceedings of the 53rd Annual Meeting of the Association for Computational Linguistics and the 7th International Joint Conference on Natural Language Processing (Volume 1: Long Papers)}, pages 451--461, Beijing, China. Association for Computational Linguistics.

\bibitem[{Eberhard et~al.(2021)Eberhard, Gary, and (eds)}]{ethnologue}
David~M Eberhard, F~Simons Gary, and Charles D~Fennig (eds). 2021.
\newblock \href {http://www.ethnologue.com.ezproxy.library.ubc.ca} {Ethnologue: Languages of the world}.
\newblock \emph{Twenty-fourth edition}, Dallas, Texas: SIL International.

\bibitem[{Ettinger et~al.(2018)Ettinger, Elgohary, Phillips, and Resnik}]{ettinger-etal-2018-assessing}
Allyson Ettinger, Ahmed Elgohary, Colin Phillips, and Philip Resnik. 2018.
\newblock \href {https://aclanthology.org/C18-1152} {Assessing composition in sentence vector representations}.
\newblock In \emph{Proceedings of the 27th International Conference on Computational Linguistics}, pages 1790--1801, Santa Fe, New Mexico, USA. Association for Computational Linguistics.

\bibitem[{Gehrmann et~al.(2021)Gehrmann, Adewumi, Aggarwal, Ammanamanchi, Aremu, Bosselut, Chandu, Clinciu, Das, Dhole, Du, Durmus, Du{\v{s}}ek, Emezue, Gangal, Garbacea, Hashimoto, Hou, Jernite, Jhamtani, Ji, Jolly, Kale, Kumar, Ladhak, Madaan, Maddela, Mahajan, Mahamood, Majumder, Martins, McMillan-Major, Mille, van Miltenburg, Nadeem, Narayan, Nikolaev, Niyongabo~Rubungo, Osei, Parikh, Perez-Beltrachini, Rao, Raunak, Rodriguez, Santhanam, Sedoc, Sellam, Shaikh, Shimorina, Sobrevilla~Cabezudo, Strobelt, Subramani, Xu, Yang, Yerukola, and Zhou}]{gehrmann-etal-2021-gem}
Sebastian Gehrmann, Tosin Adewumi, Karmanya Aggarwal, Pawan~Sasanka Ammanamanchi, Anuoluwapo Aremu, Antoine Bosselut, Khyathi~Raghavi Chandu, Miruna-Adriana Clinciu, Dipanjan Das, Kaustubh Dhole, Wanyu Du, Esin Durmus, Ond{\v{r}}ej Du{\v{s}}ek, Chris~Chinenye Emezue, Varun Gangal, Cristina Garbacea, Tatsunori Hashimoto, Yufang Hou, Yacine Jernite, Harsh Jhamtani, Yangfeng Ji, Shailza Jolly, Mihir Kale, Dhruv Kumar, Faisal Ladhak, Aman Madaan, Mounica Maddela, Khyati Mahajan, Saad Mahamood, Bodhisattwa~Prasad Majumder, Pedro~Henrique Martins, Angelina McMillan-Major, Simon Mille, Emiel van Miltenburg, Moin Nadeem, Shashi Narayan, Vitaly Nikolaev, Andre Niyongabo~Rubungo, Salomey Osei, Ankur Parikh, Laura Perez-Beltrachini, Niranjan~Ramesh Rao, Vikas Raunak, Juan~Diego Rodriguez, Sashank Santhanam, Jo{\~a}o Sedoc, Thibault Sellam, Samira Shaikh, Anastasia Shimorina, Marco~Antonio Sobrevilla~Cabezudo, Hendrik Strobelt, Nishant Subramani, Wei Xu, Diyi Yang, Akhila Yerukola, and Jiawei Zhou. 2021.
\newblock \href {https://doi.org/10.18653/v1/2021.gem-1.10} {The {GEM} benchmark: Natural language generation, its evaluation and metrics}.
\newblock In \emph{Proceedings of the 1st Workshop on Natural Language Generation, Evaluation, and Metrics (GEM 2021)}, pages 96--120, Online. Association for Computational Linguistics.

\bibitem[{Gehrmann et~al.(2022)Gehrmann, Bhattacharjee, Mahendiran, Wang, Papangelis, Madaan, Mcmillan-major, Shvets, Upadhyay, Bohnet, Yao, Wilie, Bhagavatula, You, Thomson, Garbacea, Wang, Deutsch, Xiong, Jin, Gkatzia, Radev, Clark, Durmus, Ladhak, Ginter, Winata, Strobelt, Hayashi, Novikova, Kanerva, Chim, Zhou, Clive, Maynez, Sedoc, Juraska, Dhole, Chandu, Beltrachini, Ribeiro, Tunstall, Zhang, Pushkarna, Creutz, White, Kale, Eddine, Daheim, Subramani, Dusek, Liang, Ammanamanchi, Zhu, Puduppully, Kriz, Shahriyar, Cardenas, Mahamood, Osei, Cahyawijaya, {\v{S}}tajner, Montella, Jolly, Mille, Hasan, Shen, Adewumi, Raunak, Raheja, Nikolaev, Tsai, Jernite, Xu, Sang, Liu, and Hou}]{gehrmann2022gemv2}
Sebastian Gehrmann, Abhik Bhattacharjee, Abinaya Mahendiran, Alex Wang, Alexandros Papangelis, Aman Madaan, Angelina Mcmillan-major, Anna Shvets, Ashish Upadhyay, Bernd Bohnet, Bingsheng Yao, Bryan Wilie, Chandra Bhagavatula, Chaobin You, Craig Thomson, Cristina Garbacea, Dakuo Wang, Daniel Deutsch, Deyi Xiong, Di~Jin, Dimitra Gkatzia, Dragomir Radev, Elizabeth Clark, Esin Durmus, Faisal Ladhak, Filip Ginter, Genta~Indra Winata, Hendrik Strobelt, Hiroaki Hayashi, Jekaterina Novikova, Jenna Kanerva, Jenny Chim, Jiawei Zhou, Jordan Clive, Joshua Maynez, Jo{\~a}o Sedoc, Juraj Juraska, Kaustubh Dhole, Khyathi~Raghavi Chandu, Laura~Perez Beltrachini, Leonardo F .~R. Ribeiro, Lewis Tunstall, Li~Zhang, Mahim Pushkarna, Mathias Creutz, Michael White, Mihir~Sanjay Kale, Moussa~Kamal Eddine, Nico Daheim, Nishant Subramani, Ondrej Dusek, Paul~Pu Liang, Pawan~Sasanka Ammanamanchi, Qi~Zhu, Ratish Puduppully, Reno Kriz, Rifat Shahriyar, Ronald Cardenas, Saad Mahamood, Salomey Osei, Samuel Cahyawijaya, Sanja {\v{S}}tajner,
  Sebastien Montella, Shailza Jolly, Simon Mille, Tahmid Hasan, Tianhao Shen, Tosin Adewumi, Vikas Raunak, Vipul Raheja, Vitaly Nikolaev, Vivian Tsai, Yacine Jernite, Ying Xu, Yisi Sang, Yixin Liu, and Yufang Hou. 2022.
\newblock \href {https://doi.org/10.18653/v1/2022.emnlp-demos.27} {{GEM}v2: Multilingual {NLG} benchmarking in a single line of code}.
\newblock In \emph{Proceedings of the 2022 Conference on Empirical Methods in Natural Language Processing: System Demonstrations}, pages 266--281, Abu Dhabi, UAE. Association for Computational Linguistics.

\bibitem[{Guzm{\'a}n et~al.(2019)Guzm{\'a}n, Chen, Ott, Pino, Lample, Koehn, Chaudhary, and Ranzato}]{guzman-etal-2019-flores}
Francisco Guzm{\'a}n, Peng-Jen Chen, Myle Ott, Juan Pino, Guillaume Lample, Philipp Koehn, Vishrav Chaudhary, and Marc{'}Aurelio Ranzato. 2019.
\newblock \href {https://doi.org/10.18653/v1/D19-1632} {The {FLORES} evaluation datasets for low-resource machine translation: {N}epali{--}{E}nglish and {S}inhala{--}{E}nglish}.
\newblock In \emph{Proceedings of the 2019 Conference on Empirical Methods in Natural Language Processing and the 9th International Joint Conference on Natural Language Processing (EMNLP-IJCNLP)}, pages 6098--6111, Hong Kong, China. Association for Computational Linguistics.

\bibitem[{Hasan et~al.(2021)Hasan, Bhattacharjee, Islam, Mubasshir, Li, Kang, Rahman, and Shahriyar}]{hasan-etal-2021-xl}
Tahmid Hasan, Abhik Bhattacharjee, Md.~Saiful Islam, Kazi Mubasshir, Yuan-Fang Li, Yong-Bin Kang, M.~Sohel Rahman, and Rifat Shahriyar. 2021.
\newblock \href {https://aclanthology.org/2021.findings-acl.413} {{XL}-sum: Large-scale multilingual abstractive summarization for 44 languages}.
\newblock In \emph{Findings of the Association for Computational Linguistics: ACL-IJCNLP 2021}, pages 4693--4703, Online. Association for Computational Linguistics.

\bibitem[{He et~al.(2022)He, Zhou, Ma, Berg-Kirkpatrick, and Neubig}]{he2022towards}
Junxian He, Chunting Zhou, Xuezhe Ma, Taylor Berg-Kirkpatrick, and Graham Neubig. 2022.
\newblock \href {https://openreview.net/forum?id=0RDcd5Axok} {Towards a unified view of parameter-efficient transfer learning}.
\newblock In \emph{International Conference on Learning Representations}.

\bibitem[{Jaggar(2017)}]{hausa_2017_jaggar}
Philip~J. Jaggar. 2017.
\newblock \href {http://www.jstor.org/stable/j.ctvckq4tj.6} {\emph{The Hausa “Grade 5” verb: Morphosyntactic preliminaries}}, 1 edition, pages 18--27. Harrassowitz Verlag.

\bibitem[{Joshi et~al.(2020)Joshi, Santy, Budhiraja, Bali, and Choudhury}]{joshi-etal-2020-state}
Pratik Joshi, Sebastin Santy, Amar Budhiraja, Kalika Bali, and Monojit Choudhury. 2020.
\newblock \href {https://doi.org/10.18653/v1/2020.acl-main.560} {The state and fate of linguistic diversity and inclusion in the {NLP} world}.
\newblock In \emph{Proceedings of the 58th Annual Meeting of the Association for Computational Linguistics}, pages 6282--6293, Online. Association for Computational Linguistics.

\bibitem[{Jude~Ogundepo et~al.(2022)Jude~Ogundepo, Oladipo, Adeyemi, Ogueji, and Lin}]{jude-ogundepo-etal-2022-afriteva}
Odunayo Jude~Ogundepo, Akintunde Oladipo, Mofetoluwa Adeyemi, Kelechi Ogueji, and Jimmy Lin. 2022.
\newblock \href {https://doi.org/10.18653/v1/2022.deeplo-1.14} {{A}fri{T}e{VA}: Extending ?small data? pretraining approaches to sequence-to-sequence models}.
\newblock In \emph{Proceedings of the Third Workshop on Deep Learning for Low-Resource Natural Language Processing}, pages 126--135, Hybrid. Association for Computational Linguistics.

\bibitem[{King et~al.(2022)King, Shen, Subramani, Weld, Beltagy, and Downey}]{king-etal-2022-dont}
Daniel King, Zejiang Shen, Nishant Subramani, Daniel~S. Weld, Iz~Beltagy, and Doug Downey. 2022.
\newblock \href {https://aclanthology.org/2022.gem-1.51} {Don{'}t say what you don{'}t know: Improving the consistency of abstractive summarization by constraining beam search}.
\newblock In \emph{Proceedings of the 2nd Workshop on Natural Language Generation, Evaluation, and Metrics (GEM)}, pages 555--571, Abu Dhabi, United Arab Emirates (Hybrid). Association for Computational Linguistics.

\bibitem[{Kreutzer et~al.(2021)Kreutzer, Caswell, Wang, Wahab, van Esch, Ulzii-Orshikh, Tapo, Subramani, Sokolov, Sikasote, Setyawan, Sarin, Samb, Sagot, Rivera, Rios, Papadimitriou, Osei, Suárez, Orife, Ogueji, Rubungo, Nguyen, Müller, Müller, Muhammad, Muhammad, Mnyakeni, Mirzakhalov, Matangira, Leong, Lawson, Kudugunta, Jernite, Jenny, Firat, Dossou, Dlamini, de~Silva, Çabuk Ballı, Biderman, Battisti, Baruwa, Bapna, Baljekar, Azime, Awokoya, Ataman, Ahia, Ahia, Agrawal, and Adeyemi}]{caswell2021quality}
Julia Kreutzer, Isaac Caswell, Lisa Wang, Ahsan Wahab, Daan van Esch, Nasanbayar Ulzii-Orshikh, Allahsera Tapo, Nishant Subramani, Artem Sokolov, Claytone Sikasote, Monang Setyawan, Supheakmungkol Sarin, Sokhar Samb, Benoît Sagot, Clara Rivera, Annette Rios, Isabel Papadimitriou, Salomey Osei, Pedro~Ortiz Suárez, Iroro Orife, Kelechi Ogueji, Andre~Niyongabo Rubungo, Toan~Q. Nguyen, Mathias Müller, André Müller, Shamsuddeen~Hassan Muhammad, Nanda Muhammad, Ayanda Mnyakeni, Jamshidbek Mirzakhalov, Tapiwanashe Matangira, Colin Leong, Nze Lawson, Sneha Kudugunta, Yacine Jernite, Mathias Jenny, Orhan Firat, Bonaventure F.~P. Dossou, Sakhile Dlamini, Nisansa de~Silva, Sakine Çabuk Ballı, Stella Biderman, Alessia Battisti, Ahmed Baruwa, Ankur Bapna, Pallavi Baljekar, Israel~Abebe Azime, Ayodele Awokoya, Duygu Ataman, Orevaoghene Ahia, Oghenefego Ahia, Sweta Agrawal, and Mofetoluwa Adeyemi. 2021.
\newblock \href {http://arxiv.org/abs/2103.12028} {Quality at a glance: An audit of web-crawled multilingual datasets}.
\newblock \emph{arXiv preprint arXiv:2103.12028}.

\bibitem[{Kudo and Richardson(2018)}]{kudo-richardson-2018-sentencepiece}
Taku Kudo and John Richardson. 2018.
\newblock \href {https://doi.org/10.18653/v1/D18-2012} {{S}entence{P}iece: A simple and language independent subword tokenizer and detokenizer for neural text processing}.
\newblock In \emph{Proceedings of the 2018 Conference on Empirical Methods in Natural Language Processing: System Demonstrations}, pages 66--71, Brussels, Belgium. Association for Computational Linguistics.

\bibitem[{Kumar et~al.(2022)Kumar, Shrotriya, Sahu, Mishra, Dabre, Puduppully, Kunchukuttan, Khapra, and Kumar}]{kumar-etal-2022-indicnlg}
Aman Kumar, Himani Shrotriya, Prachi Sahu, Amogh Mishra, Raj Dabre, Ratish Puduppully, Anoop Kunchukuttan, Mitesh~M. Khapra, and Pratyush Kumar. 2022.
\newblock \href {https://aclanthology.org/2022.emnlp-main.360} {{I}ndic{NLG} benchmark: Multilingual datasets for diverse {NLG} tasks in {I}ndic languages}.
\newblock In \emph{Proceedings of the 2022 Conference on Empirical Methods in Natural Language Processing}, pages 5363--5394, Abu Dhabi, United Arab Emirates. Association for Computational Linguistics.

\bibitem[{Li et~al.(2016)Li, van Deemter, and Lin}]{li-etal-2016-statistics}
Xiao Li, Kees van Deemter, and Chenghua Lin. 2016.
\newblock \href {https://doi.org/10.18653/v1/W16-6618} {Statistics-based lexical choice for {NLG} from quantitative information}.
\newblock In \emph{Proceedings of the 9th International Natural Language Generation conference}, pages 104--108, Edinburgh, UK. Association for Computational Linguistics.

\bibitem[{Liang et~al.(2020)Liang, Duan, Gong, Wu, Guo, Qi, Gong, Shou, Jiang, Cao, Fan, Zhang, Agrawal, Cui, Wei, Bharti, Qiao, Chen, Wu, Liu, Yang, Campos, Majumder, and Zhou}]{liang-etal-2020-xglue}
Yaobo Liang, Nan Duan, Yeyun Gong, Ning Wu, Fenfei Guo, Weizhen Qi, Ming Gong, Linjun Shou, Daxin Jiang, Guihong Cao, Xiaodong Fan, Ruofei Zhang, Rahul Agrawal, Edward Cui, Sining Wei, Taroon Bharti, Ying Qiao, Jiun-Hung Chen, Winnie Wu, Shuguang Liu, Fan Yang, Daniel Campos, Rangan Majumder, and Ming Zhou. 2020.
\newblock \href {https://doi.org/10.18653/v1/2020.emnlp-main.484} {{XGLUE}: A new benchmark dataset for cross-lingual pre-training, understanding and generation}.
\newblock In \emph{Proceedings of the 2020 Conference on Empirical Methods in Natural Language Processing (EMNLP)}, pages 6008--6018, Online. Association for Computational Linguistics.

\bibitem[{Liu et~al.(2020)Liu, Gu, Goyal, Li, Edunov, Ghazvininejad, Lewis, and Zettlemoyer}]{liu-etal-2020-multilingual-denoising}
Yinhan Liu, Jiatao Gu, Naman Goyal, Xian Li, Sergey Edunov, Marjan Ghazvininejad, Mike Lewis, and Luke Zettlemoyer. 2020.
\newblock \href {https://doi.org/10.1162/tacl_a_00343} {Multilingual denoising pre-training for neural machine translation}.
\newblock \emph{Transactions of the Association for Computational Linguistics}, 8:726--742.

\bibitem[{Marvin and Linzen(2018)}]{marvin-linzen-2018-targeted}
Rebecca Marvin and Tal Linzen. 2018.
\newblock \href {https://doi.org/10.18653/v1/D18-1151} {Targeted syntactic evaluation of language models}.
\newblock In \emph{Proceedings of the 2018 Conference on Empirical Methods in Natural Language Processing}, pages 1192--1202, Brussels, Belgium. Association for Computational Linguistics.

\bibitem[{McCoy et~al.(2019)McCoy, Pavlick, and Linzen}]{mccoy-etal-2019-right}
Tom McCoy, Ellie Pavlick, and Tal Linzen. 2019.
\newblock \href {https://doi.org/10.18653/v1/P19-1334} {Right for the wrong reasons: Diagnosing syntactic heuristics in natural language inference}.
\newblock In \emph{Proceedings of the 57th Annual Meeting of the Association for Computational Linguistics}, pages 3428--3448, Florence, Italy. Association for Computational Linguistics.

\bibitem[{Muennighoff et~al.(2022)Muennighoff, Wang, Sutawika, Roberts, Biderman, Scao, Bari, Shen, Yong, Schoelkopf, Tang, Radev, Aji, Almubarak, Albanie, Alyafeai, Webson, Raff, and Raffel}]{muennighoff2022crosslingual}
Niklas Muennighoff, Thomas Wang, Lintang Sutawika, Adam Roberts, Stella Biderman, Teven~Le Scao, M~Saiful Bari, Sheng Shen, Zheng-Xin Yong, Hailey Schoelkopf, Xiangru Tang, Dragomir Radev, Alham~Fikri Aji, Khalid Almubarak, Samuel Albanie, Zaid Alyafeai, Albert Webson, Edward Raff, and Colin Raffel. 2022.
\newblock \href {http://arxiv.org/abs/2211.01786} {Crosslingual generalization through multitask finetuning}.

\bibitem[{Nallapati et~al.(2016)Nallapati, Zhou, dos Santos, Gul{\c{c}}ehre, and Xiang}]{nallapati-etal-2016-abstractive}
Ramesh Nallapati, Bowen Zhou, Cicero dos Santos, {\c{C}}a{\u{g}}lar Gul{\c{c}}ehre, and Bing Xiang. 2016.
\newblock \href {https://doi.org/10.18653/v1/K16-1028} {Abstractive text summarization using sequence-to-sequence {RNN}s and beyond}.
\newblock In \emph{Proceedings of the 20th {SIGNLL} Conference on Computational Natural Language Learning}, pages 280--290, Berlin, Germany. Association for Computational Linguistics.

\bibitem[{Nzeyimana and Niyongabo~Rubungo(2022)}]{nzeyimana-niyongabo-rubungo-2022-kinyabert}
Antoine Nzeyimana and Andre Niyongabo~Rubungo. 2022.
\newblock \href {https://doi.org/10.18653/v1/2022.acl-long.367} {{K}inya{BERT}: a morphology-aware {K}inyarwanda language model}.
\newblock In \emph{Proceedings of the 60th Annual Meeting of the Association for Computational Linguistics (Volume 1: Long Papers)}, pages 5347--5363, Dublin, Ireland. Association for Computational Linguistics.

\bibitem[{Ogueji and Ahia(2019)}]{ogueji2019pidginunmt}
Kelechi Ogueji and Orevaoghene Ahia. 2019.
\newblock Pidginunmt: Unsupervised neural machine translation from west african pidgin to english.
\newblock \emph{arXiv preprint arXiv:1912.03444}.

\bibitem[{Ogueji et~al.(2021)Ogueji, Zhu, and Lin}]{ogueji-etal-2021-small}
Kelechi Ogueji, Yuxin Zhu, and Jimmy Lin. 2021.
\newblock \href {https://doi.org/10.18653/v1/2021.mrl-1.11} {Small data? no problem! exploring the viability of pretrained multilingual language models for low-resourced languages}.
\newblock In \emph{Proceedings of the 1st Workshop on Multilingual Representation Learning}, pages 116--126, Punta Cana, Dominican Republic. Association for Computational Linguistics.

\bibitem[{Palivela(2021)}]{PALIVELA2021100025}
Hemant Palivela. 2021.
\newblock \href {https://doi.org/https://doi.org/10.1016/j.jjimei.2021.100025} {Optimization of paraphrase generation and identification using language models in natural language processing}.
\newblock \emph{International Journal of Information Management Data Insights}, 1(2):100025.

\bibitem[{Radford et~al.(2018)Radford, Narasimhan, Salimans, and Sutskever}]{radford2018improving}
Alec Radford, Karthik Narasimhan, Tim Salimans, and Ilya Sutskever. 2018.
\newblock Improving language understanding by generative pre-training.

\bibitem[{Radford et~al.(2019)Radford, Wu, Child, Luan, Amodei, Sutskever et~al.}]{radford2019language}
Alec Radford, Jeffrey Wu, Rewon Child, David Luan, Dario Amodei, Ilya Sutskever, et~al. 2019.
\newblock Language models are unsupervised multitask learners.
\newblock \emph{OpenAI blog}, 1(8):9.

\bibitem[{Raffel et~al.(2020)Raffel, Shazeer, Roberts, Lee, Narang, Matena, Zhou, Li, and Liu}]{JMLR:v21:20-074}
Colin Raffel, Noam Shazeer, Adam Roberts, Katherine Lee, Sharan Narang, Michael Matena, Yanqi Zhou, Wei Li, and Peter~J. Liu. 2020.
\newblock \href {http://jmlr.org/papers/v21/20-074.html} {Exploring the limits of transfer learning with a unified text-to-text transformer}.
\newblock \emph{Journal of Machine Learning Research}, 21(140):1--67.

\bibitem[{Reid et~al.(2021{\natexlab{a}})Reid, Hu, Neubig, and Matsuo}]{reid21afromt}
Machel Reid, Junjie Hu, Graham Neubig, and Yutaka Matsuo. 2021{\natexlab{a}}.
\newblock \href {https://arxiv.org/abs/2109.04715} {Afro{MT}: Pretraining strategies and reproducible benchmarks for translation of 8 african languages}.
\newblock In \emph{Conference on Empirical Methods in Natural Language Processing (EMNLP)}, Punta Cana, Dominican Republic.

\bibitem[{Reid et~al.(2021{\natexlab{b}})Reid, Hu, Neubig, and Matsuo}]{reid-etal-2021-afromt}
Machel Reid, Junjie Hu, Graham Neubig, and Yutaka Matsuo. 2021{\natexlab{b}}.
\newblock \href {https://doi.org/10.18653/v1/2021.emnlp-main.99} {{A}fro{MT}: Pretraining strategies and reproducible benchmarks for translation of 8 {A}frican languages}.
\newblock In \emph{Proceedings of the 2021 Conference on Empirical Methods in Natural Language Processing}, pages 1306--1320, Online and Punta Cana, Dominican Republic. Association for Computational Linguistics.

\bibitem[{Ruder et~al.(2019)Ruder, Peters, Swayamdipta, and Wolf}]{ruder-etal-2019-transfer}
Sebastian Ruder, Matthew~E. Peters, Swabha Swayamdipta, and Thomas Wolf. 2019.
\newblock \href {https://doi.org/10.18653/v1/N19-5004} {Transfer learning in natural language processing}.
\newblock In \emph{Proceedings of the 2019 Conference of the North {A}merican Chapter of the Association for Computational Linguistics: Tutorials}, pages 15--18, Minneapolis, Minnesota. Association for Computational Linguistics.

\bibitem[{Scherrer(2020)}]{scherrer_yves_2020_3707949}
Yves Scherrer. 2020.
\newblock \href {https://aclanthology.org/2020.lrec-1.848} {{T}a{P}a{C}o: A corpus of sentential paraphrases for 73 languages}.
\newblock In \emph{Proceedings of the Twelfth Language Resources and Evaluation Conference}, pages 6868--6873, Marseille, France. European Language Resources Association.

\bibitem[{Sennrich et~al.(2016)Sennrich, Haddow, and Birch}]{sennrich-etal-2016-neural}
Rico Sennrich, Barry Haddow, and Alexandra Birch. 2016.
\newblock \href {https://doi.org/10.18653/v1/P16-1162} {Neural machine translation of rare words with subword units}.
\newblock In \emph{Proceedings of the 54th Annual Meeting of the Association for Computational Linguistics (Volume 1: Long Papers)}, pages 1715--1725, Berlin, Germany. Association for Computational Linguistics.

\bibitem[{Shi et~al.(2021)Shi, Keneshloo, Ramakrishnan, and Reddy}]{shi_2021}
Tian Shi, Yaser Keneshloo, Naren Ramakrishnan, and Chandan~K. Reddy. 2021.
\newblock \href {https://doi.org/10.1145/3419106} {Neural abstractive text summarization with sequence-to-sequence models}.
\newblock \emph{ACM/IMS Trans. Data Sci.}, 2(1).

\bibitem[{Sutskever et~al.(2014)Sutskever, Vinyals, and Le}]{sutskever_2014}
Ilya Sutskever, Oriol Vinyals, and Quoc~V. Le. 2014.
\newblock Sequence to sequence learning with neural networks.
\newblock In \emph{Proceedings of the 27th International Conference on Neural Information Processing Systems - Volume 2}, NIPS'14, page 3104–3112. MIT Press.

\bibitem[{Van~Deemter et~al.(2005)Van~Deemter, Krahmer, and Theune}]{deemter_2005}
Kees Van~Deemter, Emiel Krahmer, and Mari\"{e}t Theune. 2005.
\newblock \href {https://doi.org/10.1162/0891201053630291} {Real versus template-based natural language generation: A false opposition?}
\newblock \emph{Comput. Linguist.}, 31(1):15–24.

\bibitem[{van Miltenburg et~al.(2020)van Miltenburg, van~der Lee, Castro-Ferreira, and Krahmer}]{van-miltenburg-etal-2020-evaluation}
Emiel van Miltenburg, Chris van~der Lee, Thiago Castro-Ferreira, and Emiel Krahmer. 2020.
\newblock \href {https://aclanthology.org/2020.evalnlgeval-1.3} {Evaluation rules! on the use of grammars and rule-based systems for {NLG} evaluation}.
\newblock In \emph{Proceedings of the 1st Workshop on Evaluating NLG Evaluation}, pages 17--27, Online (Dublin, Ireland). Association for Computational Linguistics.

\bibitem[{Vaswani et~al.(2017)Vaswani, Shazeer, Parmar, Uszkoreit, Jones, Gomez, Kaiser, and Polosukhin}]{vaswani_2017}
Ashish Vaswani, Noam Shazeer, Niki Parmar, Jakob Uszkoreit, Llion Jones, Aidan~N. Gomez, \L{}ukasz Kaiser, and Illia Polosukhin. 2017.
\newblock Attention is all you need.
\newblock In \emph{Proceedings of the 31st International Conference on Neural Information Processing Systems}, NIPS'17, page 6000–6010, Red Hook, NY, USA. Curran Associates Inc.

\bibitem[{Xue et~al.(2021)Xue, Constant, Roberts, Kale, Al-Rfou, Siddhant, Barua, and Raffel}]{xue-etal-2021-mt5}
Linting Xue, Noah Constant, Adam Roberts, Mihir Kale, Rami Al-Rfou, Aditya Siddhant, Aditya Barua, and Colin Raffel. 2021.
\newblock \href {https://doi.org/10.18653/v1/2021.naacl-main.41} {m{T}5: A massively multilingual pre-trained text-to-text transformer}.
\newblock In \emph{Proceedings of the 2021 Conference of the North American Chapter of the Association for Computational Linguistics: Human Language Technologies}, pages 483--498, Online. Association for Computational Linguistics.

\end{thebibliography}
\bibliographystyle{acl_natbib}
\appendix 
\clearpage
\appendixpage
\addappheadtotoc
\numberwithin{figure}{section}
\numberwithin{table}{section}


\section{Pretraining Data}\label{app:pretraing-data}
We provide details of our pretraining data below:
\noindent{\textbf{Religious Domain.}}
Our religious data is taken from online Bibles, Qurans, and data crawled from the Jehovah's witness website. We also include religious texts from the book of Mormon. 

\noindent{\textbf{News Domain.}}
We collect data from online newspapers~\cite{adebara-abdul-mageed-2022-towards} and news sites such as Voice of America, Voice of Nigeria, BBC, Global voices, and DW news sites. We collect local newspapers from $27$ languages from across Africa. 

\noindent{\textbf{Government Documents.}}
We collect government documents \href{https://www.sadilar.org/index.php/en/}{South African Centre for Digital Language Resources} (SADiLaR), and the \href{https://www.ohchr.org/en/udhr/pages/searchbylang.aspx}{Universal Declaration of human rights} (UDHR) in multiple languages.

\noindent{\textbf{Health Documents.}}
We collect multiple health documents from the Department of Health, State Government of Victoria, Australia. We collect documents in Amharic, Dinka, Harari, Oromo, Somali, Swahili, and Tigrinya.

\noindent{\textbf{Existing Corpora.}}
We collect corpora available on the web for different African languages, including from \href{https://www.gutenberg.org/browse/languages/af}{Project Gutenberg} for Afrikaans, \href{https://zenodo.org/record/3668495#.YcTXu2DMJyy}{South African News data.} for Sepedi and Setswana, OSCAR \cite{AbadjiOrtizSuarezRomaryetal.2021} for Afrikaans, Amharic, Somali, Swahili, Oromo, Malagasy, and Yoruba. We also used \href{https://opus.nlpl.eu/Tatoeba.php}{Tatoeba} for Afrikaans, Amharic, Bemba, Igbo, Kanuri, Kongo, Luganda, Malagasy, Sepedi, Ndebele, Kinyarwanda, Somali, Swahili, Tsonga, Xhosa, Yoruba, and Zulu; \href{https://zenodo.org/record/3553423#.YcTXkWDMJyx}{Swahili Language Modelling Data} for Swahili; \href{https://github.com/ijdutse/hausa-corpus/blob/master/data/all-merged-hausa-datasets.txt}{Ijdutse corpus} for  Hausa; \href{https://github.com/AI-Lab-Makerere/Data4Good}{Data4Good corpora} for Luganda, CC-100 for Amharic, Fulah, Igbo, Yoruba, Hausa, Tswana, Lingala, Luganada, Afrikaans, Somali, Swahili, Swati, North Sotho, Oromo, Wolof, Xhosa, and Zulu; \href{https://huggingface.co/datasets/castorini/afriberta-corpus}{Afriberta-Corpus} for Afaan / Oromo, Amharic, Gahuza, Hausa, Igbo, Pidgin, Somali, Swahili, Tigrinya and Yoruba; \href{https://huggingface.co/datasets/mc4}{mC4} for Afrikaans, Amharic, Hausa, Igbo, Malagasy, Chichewa, Shona, Somali, Sepedi, Swahili, Xhosa, Yoruba and Zulu. 

\section{AfroNLG Benchmark}\label{app_sec:benchmark}
We report statistics of AfroNLG benchmark in Table \ref{tab:benchmarkmt} and \ref{tab:benchmarkcompare} respectively.
\begin{table}[h!]
\scriptsize
\centering
\resizebox{0.9\columnwidth}{!}{%
\begin{tabular}{clccccc}
\toprule  \textbf{Dataset}&\textbf{Pairs}                                                                                                                                                                                                                                                   & \textbf{Train} & \textbf{Dev} & \textbf{Test}            \\ \toprule
\multirow{15}{*}{\rotatebox[origin=]{90}{\textbf{Lafand} }    } & eng-hau                                                                                                                                                                                                                                                                     & 5,866          & 1,301          & 1,501                    \\ 
                    &                                                        eng-ibo                                                                                                                                                                                                                                                                     & 6,945          & 1,457          & 1,412                    \\
                    &                                                     eng-lug                                                                                                                                                                                                                                                                     & 4,076          & 1,501          & 1,501                    \\
                    &                                                       eng-pcm                                                                                                                                                                                                                                                                     & 4,791          & 1,485          & 1,565                    \\
                    &                                                      eng-swa                                                                                                                                                                                                                                                                     & 30,783         & 1,792          & 1,836                    \\
                    &                                                      eng-tsn                                                                                                                                                                                                                                                                     & 2,101          & 1,343          & 1,501                    \\
                    &                                                    eng-twi                                                                                                                                                                                                                                                                     & 3,338          & 1,285          & 1,501                    \\
                    &                                                    eng-yor                                                                                                                                                                                                                                                                     & 6,645          & 1,545          & 1,559                    \\
                    &                                                   eng-zul                                                                                                                                                                                                                                                                     & 3,540          & 1,462          & 1,001                    \\
                    &                                                    fra-bam                                                                                                                                                                                                                                                                     & 3,014          & 1,501          & 1,501                    \\
                    &                                                  fra-bbj                                                                                                                                                                                                                                                                     & 2,233          & 1,134          & 1,431                    \\
                    &                                                     fra-ewe                                                                                                                                                                                                                                                                     & 2,027          & 1,415          & 1,564                    \\
                    &                                                    fra-fon                                                                                                                                                                                                                                                                     & 2,638          & 1,228          & 1,580                    \\
                    &                                                     fra-mos                                                                                                                                                                                                                                                                     & 2,494          & 1,493          & 1,575                    \\
                    &                                                     fra-wol                                                                                                                                                                                                                                                                     & 3,361          & 1,507          & 1,501                    \\ \hline
                   \multirow{8}{*}{\rotatebox[origin=]{90}{\textbf{AfroMT} }    }    &                                        eng-afr                                                                                                                                                                                                                                                                     & 25,799         & 3,226          & 3,226                    \\
                    &                                                    eng-bem                                                                                                                                                                                                                                                                     & 12,043         & 1,506          & 1,506                    \\
                    &                                               eng-lin                                                                                                                                                                                                                                                                     & 17,679         & 2,211          & 2,210                    \\
                    &                                               eng-run                                                                                                                                                                                                                                                                     & 12,475         & 1,560          & 1,560                    \\
                    &                                            eng-sot                                                                                                                                                                                                                                                                     & 28,844         & 3,607          & 3,606                    \\
                    &                                                 eng-swa                                                                                                                                                                                                                                                                     & 28,084         & 3,511          & 3,512                    \\
                    &                                                  eng-xho                                                                                                                                                                                                                                                                     & 26,091         & 3,263          & 3,262                    \\
                    &                                                    eng-zul                                                                                                                                                                                                                                                                     & 29,127         & 3,641          & 3,642                    \\ \hline
                  \textbf{PidginUNMT}     &                                    eng-pcm                                                                                                                                                                                                                                                                     & 1,682          & 211            & 211                      \\\hline
                   \textbf{SALT}    &             All-pairs   & 20,006         & 2,501          & 2,502                    \\
 \bottomrule       
\end{tabular}
}
\caption{Statistics of the MT data in our benchmark. All-pairs each have the same size of data. They include ach-eng, ach-lgg, ach-lug, ach-nyn, ach-teo, ach-teo, eng-lgg, eng-lug, eng-nyn, eng-teo, lgg-teo, lug-lgg, lug-teo, nyn-lgg, nyn-lug, and nyn-teo  }
\label{tab:benchmarkmt}
\end{table}

\begin{table*}[]
\scriptsize
\centering
\resizebox{1.6\columnwidth}{!}{%
\begin{tabular}{clllccccc}
\toprule
\textbf{Task Cluster} & \textbf{Test Set} & \textbf{Source} & \textbf{Train} & \textbf{Dev} & \textbf{Test} \\ \toprule
Cloze test & 517 languages & Ours & 103,400  & 25,850 & 51,700 \\ \hline
Paraphrase & Multilingual\textsuperscript{$\dagger\dagger$} & \cite{scherrer_yves_2020_3707949} & 22,390 & 2,797 & 2,794 \\
 & Berber & & 17,607 & 2,200 & 2,200 \\  
 & Kabyle & & 4,441 & 555 & 555 \\  \midrule
Question Answering & Swahili & \cite{clark-etal-2020-tydi} & 49,881 & 499 & n/a \\  \midrule
\multirow{13}{*}{\rotatebox[origin=]{90}{Summarization}} & Multilingual\textsuperscript{$\dagger$} & \cite{hasan-etal-2021-xl} & 63,040 & 7,875 & 7875 \\
 & Amharic & & 5,761 & 719 & 719 \\  
 & Igbo & & 4,183 & 522 & 522 \\  
 & Oromo & & 6,063 & 757 & 757 \\  
 & Rundi & & 5,746 & 718 & 718 \\  
 & Swahili & & 7,898 & 987 & 987 \\  
 & Yor\`{u}b\'{a} & & 6,350 & 793 & 793 \\  
 & Hausa & & 6,418 & 802 & 802 \\  
 & Nigerian Pidgin & & 9,208 & 1,151 & 1,151 \\  
 & Somali & & 5,962 & 745 & 745 \\  
 & Tigrinya & & 5,451 & 681 & 681 \\ 
 & Multilingual\textsuperscript{$\star\dagger$} & Ours & & & 428 \\ \hline
\multirow{13}{*}{\rotatebox[origin=]{90}{Title Generation}} & Multilingual\textsuperscript{$\dagger$} & \cite{hasan-etal-2021-xl} & 63,040 & 7,875 & 7875 \\
 & Amharic & & 5,761 & 719 & 719 \\  
 & Igbo & & 4,183 & 522 & 522 \\  
 & Oromo & & 6,063 & 757 & 757 \\  
 & Rundi & & 5,746 & 718 & 718 \\  
 & Swahili & & 7,898 & 987 & 987 \\  
 & Yor\`{u}b\'{a} & & 6,350 & 793 & 793 \\  
 & Hausa & & 6,418 & 802 & 802 \\  
 & Nigerian Pidgin & & 9,208 & 1,151 & 1,151 \\  
 & Somali & & 5,962 & 745 & 745 \\  
 & Tigrinya & & 5,451 & 681 & 681 \\
 & Multilingual\textsuperscript{$\star$} & Ours & & & 5899 \\ 
\bottomrule       
\end{tabular}
}
\caption{Statistics of the data in our benchmark. \textsuperscript{$\dagger\dagger$} includes amh, ber, kab, run. \textsuperscript{$\dagger$} has amh, ibo, orm, run, swa, yor, hau, pcm, som, and tir. \textsuperscript{$\star\dagger$} is a newly created summarization test set including `hau', `nde' (zero-shot), and `swa'. \textsuperscript{$\star$} is a newly created test set across 15 languages: `amh', `gag' (zero-shot), `hau', `ibo', `pcm', `som', `swa', `tir', `yor', `kin' (zero-shot), `afr', `mlg' (zero-shot), `orm', `nde' (zero-shot), `sna'(zero-shot)}
\label{tab:benchmarkothers}
\end{table*}

\subsection{CHRF and CHRF++ Results}
\begin{table*}[!]
\scriptsize
\centering
\begin{tabular}{lllllll} \toprule
 \textbf{Task}                                              & \textbf{Metric} & \textbf{mT0}         &\textbf{ mT5 }       & \textbf{afri-mt5 }  & \textbf{AfriTeVa}    & \textbf{Cheetah }   \\ \toprule
Translate English to Afrikaans                    & Chrf   & 26.97\textsuperscript{$\pm$4.75} & 26.11\textsuperscript{$\pm$4.12}  & 14.66\textsuperscript{$\pm$8.79}  & 20.75\textsuperscript{$\pm$4.02}  & \textbf{39.88\textsuperscript{$\pm$0.81}} \\
        Translate English to Bemba                        & Chrf   & 10.27\textsuperscript{$\pm$0.89} & 6.39\textsuperscript{$\pm$1.96}   & 20.23\textsuperscript{$\pm$13.97} & 9.94\textsuperscript{$\pm$10.05}  & \textbf{15.76\textsuperscript{$\pm$0.19}} \\
        Translate English to Rundi                        & Chrf   & 21.51\textsuperscript{$\pm$1.39} & 17.56\textsuperscript{$\pm$3.13}  & 24.91\textsuperscript{$\pm$3.59}  & \textbf{31.58\textsuperscript{$\pm$2.33}}  & 28.65\textsuperscript{$\pm$3.55} \\
        Translate English to Sesotho                      & Chrf   & 21.08\textsuperscript{$\pm$3.54} & 12.08\textsuperscript{$\pm$10.91} & 23.75\textsuperscript{$\pm$4.77}  & \textbf{29.57\textsuperscript{$\pm$1.61}}  & 29.05\textsuperscript{$\pm$2.41} \\
        Translate English to Swahili                      & Chrf   & 23.26\textsuperscript{$\pm$0.16} & 20.35\textsuperscript{$\pm$4.87}  & 24.60\textsuperscript{$\pm$0.2}   & 20.5\textsuperscript{$\pm$4.88}   & \textbf{37.24\textsuperscript{$\pm$0.04}} \\
        Translate English to Xhosa                        & Chrf   & 27.44\textsuperscript{$\pm$3.1}  & 25.88\textsuperscript{$\pm$4.94}  & \textbf{34.97\textsuperscript{$\pm$2.49}}  & 20.25\textsuperscript{$\pm$15.35} & 33.45\textsuperscript{$\pm$0.21} \\
        Translate English to Zulu                         & Chrf   & 27.12\textsuperscript{$\pm$3.49} & 21.54\textsuperscript{$\pm$2.16}  & 37.8\textsuperscript{$\pm$1.41}   & 25.39\textsuperscript{$\pm$16.55} & \textbf{43.75\textsuperscript{$\pm$0.11}} \\
        Translate English to Hausa                        & Chrf   & 28.53\textsuperscript{$\pm$0.26} & 27.65\textsuperscript{$\pm$0.53}  & 19.99\textsuperscript{$\pm$0.42}  & 31.68\textsuperscript{$\pm$0.29}  & \textbf{34.9\textsuperscript{$\pm$0.32}} \\
        Translate English to Igbo                         & Chrf   & 40.31\textsuperscript{$\pm$0.17} & 37.18\textsuperscript{$\pm$0.34}  & 22.01\textsuperscript{$\pm$0.7}  & 33.24\textsuperscript{$\pm$0.23}  & \textbf{44.37\textsuperscript{$\pm$0.31}} \\
        Translate English to Luganda                      & Chrf   & 25.94\textsuperscript{$\pm$2.41} & 23.33\textsuperscript{$\pm$0.31}  & 15.57\textsuperscript{$\pm$1.45}  & 24.16\textsuperscript{$\pm$2.55}  & \textbf{36.22\textsuperscript{$\pm$0.09}} \\
        Translate English to N. Pidgin              & Chrf   & 63.49\textsuperscript{$\pm$0.05} & \textbf{63.9\textsuperscript{$\pm$0.1}}    & 24.79\textsuperscript{$\pm$0.68}  & 53.76\textsuperscript{$\pm$0.01}  & 62.95\textsuperscript{$\pm$0.17} \\
        Translate English to Swahili                      & Chrf   & 50.52\textsuperscript{$\pm$3.33} & 51.76\textsuperscript{$\pm$0.12}  & 21.00\textsuperscript{$\pm$0.7}  & 44.84\textsuperscript{$\pm$0.33}  & \textbf{56.36\textsuperscript{$\pm$0.15}} \\
        Translate English to Setswana                     & Chrf   & 30.89\textsuperscript{$\pm$0.36} & 16.62\textsuperscript{$\pm$0.22}  & 13.17\textsuperscript{$\pm$1.73}  & 23.75\textsuperscript{$\pm$0.45}  & \textbf{35.87\textsuperscript{$\pm$0.64}} \\
        Translate English to Twi                          & Chrf   & 23.56\textsuperscript{$\pm$0.24} & 15.8\textsuperscript{$\pm$1.29}   & 12.74\textsuperscript{$\pm$1.33}  & 17.47\textsuperscript{$\pm$3.26}  & \textbf{25.89\textsuperscript{$\pm$0.2}} \\
        Translate English to Yoruba                       & Chrf   & 19.41\textsuperscript{$\pm$1.97} & 16.51\textsuperscript{$\pm$0.38}  & 11.49\textsuperscript{$\pm$0.29}  & 20.62\textsuperscript{$\pm$0.36}  & \textbf{25.09\textsuperscript{$\pm$0.07}} \\
        Translate English to Zulu                         & Chrf   & 35.4\textsuperscript{$\pm$1.27}  & 16.13\textsuperscript{$\pm$7.84}  & 15.04\textsuperscript{$\pm$1.1}   & 12.75\textsuperscript{$\pm$0.56}  & \textbf{38.81\textsuperscript{$\pm$0.21}} \\
        Translate French to Bambara                       & Chrf   & 16.49\textsuperscript{$\pm$0.39} & 7.44\textsuperscript{$\pm$1.12}   & 10.16\textsuperscript{$\pm$1.58}  & 19.41\textsuperscript{$\pm$0.53}  & \textbf{19.91\textsuperscript{$\pm$0.05}} \\
        Translate French to Ghomálá’                      & Chrf   & 8.3\textsuperscript{$\pm$0.76}   & 6.53\textsuperscript{$\pm$0.57}   & 6.72\textsuperscript{$\pm$3.75}   & \textbf{13.16\textsuperscript{$\pm$0.4}}  & 8.57\textsuperscript{$\pm$3.15}  \\
        Translate French to Ewe                           & Chrf   & 10.19\textsuperscript{$\pm$2.32} & 5.46\textsuperscript{$\pm$3.02}   & 6.96\textsuperscript{$\pm$3.02}   & 13.44\textsuperscript{$\pm$1.64}  & \textbf{21.6\textsuperscript{$\pm$0.22}}  \\
        Translate French to Fon                           & Chrf   & 5.67\textsuperscript{$\pm$2.65}  & 6.09\textsuperscript{$\pm$0.72}   & 5.82\textsuperscript{$\pm$1.58}   & 11.88\textsuperscript{$\pm$1.83}  & \textbf{12.71\textsuperscript{$\pm$0.41}} \\
        Translate French to Moore                         & Chrf   & 7.86\textsuperscript{$\pm$1.43}  & 5.16\textsuperscript{$\pm$2.20}   & 7.79\textsuperscript{$\pm$0.97}   & 11.42\textsuperscript{$\pm$0.7}   & \textbf{12.34\textsuperscript{$\pm$0.56}} \\
        Translate French to Wolof                         & Chrf   & 17.55\textsuperscript{$\pm$0.2}  & 3.15\textsuperscript{$\pm$0.12}   & 11.26\textsuperscript{$\pm$1.91}  & \textbf{17.58\textsuperscript{$\pm$0.44} } & 16.67\textsuperscript{$\pm$0.21} \\
        Translate English to N. Pidgin (pidginUNMT) & Chrf   & 41.83\textsuperscript{$\pm$0.17} & 37.12\textsuperscript{$\pm$0.77}  & 21.65\textsuperscript{$\pm$1.33}  & 39.04\textsuperscript{$\pm$0.50} & \textbf{40.2\textsuperscript{$\pm$0.17}}  \\
        Translate Acholi to English                       & Chrf   & 39.12\textsuperscript{$\pm$0.1}  & 33.07\textsuperscript{$\pm$5.49}  & 21.65\textsuperscript{$\pm$1.33}  & 34.19\textsuperscript{$\pm$0.06}  & \textbf{42.17\textsuperscript{$\pm$0.05}} \\
        Translate Acholi to Lugbara                       & Chrf   & 25.05\textsuperscript{$\pm$0.85} & 20.61\textsuperscript{$\pm$5.92}  & 28.71\textsuperscript{$\pm$0.34}  & \textbf{34.01\textsuperscript{$\pm$0.29}}  & 32.31\textsuperscript{$\pm$1.11} \\
        Translate Acholi to Luganda                       & Chrf   & 22.13\textsuperscript{$\pm$0.63} & 25.75\textsuperscript{$\pm$0.02}  & 24.31\textsuperscript{$\pm$0.1}   & 32.77\textsuperscript{$\pm$0.68}  & \textbf{37.34\textsuperscript{$\pm$0.47}} \\
        Translate Acholi to Nyankore                      & Chrf   & 27.52\textsuperscript{$\pm$0.45} & 20.03\textsuperscript{$\pm$3.88}  & 24.50\textsuperscript{$\pm$0.02}  & 32.39\textsuperscript{$\pm$0.92}  & \textbf{35.0\textsuperscript{$\pm$0.33}}  \\
        Translate Acholi to Ateso                         & Chrf   & 26.0\textsuperscript{$\pm$1.99}  & 22.16\textsuperscript{$\pm$1.63}  & 28.33\textsuperscript{$\pm$0.01}  & \textbf{35.37\textsuperscript{$\pm$0.61}}  & 34.62\textsuperscript{$\pm$1.05} \\
        Translate English to Lugbara                      & Chrf   & 38.84\textsuperscript{$\pm$0.01} & 37.12\textsuperscript{$\pm$0.77}  & 39.11\textsuperscript{$\pm$0.01}  & 38.94\textsuperscript{$\pm$0.3}   & \textbf{40.2\textsuperscript{$\pm$0.17}}  \\
        Translate English to Luganda                      & Chrf   & 43.71\textsuperscript{$\pm$0.08} & 41.05\textsuperscript{$\pm$0.19}  & 35.34\textsuperscript{$\pm$1.11}  & 43.14\textsuperscript{$\pm$0.22}  & \textbf{49.38\textsuperscript{$\pm$0.02}} \\
        Translate English to Nyankore                     & Chrf   & 40.43\textsuperscript{$\pm$0.21} & 38.38\textsuperscript{$\pm$0.13}  & 36.8\textsuperscript{$\pm$0.07}   & 40.36\textsuperscript{$\pm$0.17}  & \textbf{43.67\textsuperscript{$\pm$0.32}} \\
        Translate English to Ateso (salt)                 & Chrf   & 41.98\textsuperscript{$\pm$0.13} & 38.91\textsuperscript{$\pm$0.05}  & 39.76\textsuperscript{$\pm$1.35}  & 42.1\textsuperscript{$\pm$0.42}   & \textbf{42.96\textsuperscript{$\pm$0.48}} \\
        Translate Lugbara to Ateso                        & Chrf   & 22.67\textsuperscript{$\pm$1.51} & 20.47\textsuperscript{$\pm$0.7}  & 28.13\textsuperscript{$\pm$0.58}  & \textbf{34.3\textsuperscript{$\pm$0.64}}   & 29.04\textsuperscript{$\pm$0.3}  \\
        Translate Luganda to Lugbara                      & Chrf   & 28.65\textsuperscript{$\pm$1.5}  & 25.74\textsuperscript{$\pm$0.5}   & 30.87\textsuperscript{$\pm$0.12}  & 34.26\textsuperscript{$\pm$0.24} & \textbf{34.94\textsuperscript{$\pm$0.6}}  \\
        Translate Luganda to Ateso                        & Chrf   & 31.74\textsuperscript{$\pm$0.22} & 27.66\textsuperscript{$\pm$0.64}  & 34.04\textsuperscript{$\pm$0.01}  & 37.19\textsuperscript{$\pm$0.07}  & \textbf{39.05\textsuperscript{$\pm$0.49}} \\
        Translate Nyankore to Lugbara                     & Chrf   & 27.47\textsuperscript{$\pm$0.45} & 24.63\textsuperscript{$\pm$0.76}  & 15.01\textsuperscript{$\pm$0.01}  & \textbf{33.17\textsuperscript{$\pm$0.21}}  & 33.2\textsuperscript{$\pm$0.19} \\
        Translate Nyankore to Luganda                     & Chrf   & 39.34\textsuperscript{$\pm$0.14} & 37.34\textsuperscript{$\pm$0.16}  & 35.26\textsuperscript{$\pm$0.13}  & 40.48\textsuperscript{$\pm$0.63}  & \textbf{45.29\textsuperscript{$\pm$0.01}} \\
        Translate Nyankore to Ateso                       & Chrf   & 28.6\textsuperscript{$\pm$0.11}  & 24.64\textsuperscript{$\pm$1.05}  & 30.69\textsuperscript{$\pm$0.16}  & 34.37\textsuperscript{$\pm$0.14}  & \textbf{35.52\textsuperscript{$\pm$0.64}} \\ \bottomrule
        &\textbf{Average}&28.07	
        &23.88
        &22.62	
        &28.77	&\textbf{34.08}\\
        \bottomrule
\end{tabular}\caption{Performance of various models on MT data using CHRF}\label{tab:results_chrf}
\end{table*}
\begin{table*}[!]
\scriptsize
\centering
\begin{tabular}{lllllll} \toprule
\textbf{Task}                                              & \textbf{Metric} & \textbf{mT0}         &\textbf{ mT5 }       & \textbf{afri-mt5 }  & \textbf{AfriTeVa}    & \textbf{Cheetah }   \\ \toprule

Translate English to Afrikaans                    & Chrf++ & 22.86\textsuperscript{$\pm$3.74}  & 22.32\textsuperscript{$\pm$2.80} & 11.62\textsuperscript{$\pm$6.72} & 17.27\textsuperscript{$\pm$2.91}  & \textbf{34.02\textsuperscript{$\pm$0.7}}  \\
        Translate English to Bemba                        & Chrf++ & 9.04\textsuperscript{$\pm$0.79}   & 5.46\textsuperscript{$\pm$1.78}  & \textbf{23.65\textsuperscript{$\pm$1.87}} & 7.85\textsuperscript{$\pm$7.45}   & 13.9\textsuperscript{$\pm$0.13}  \\
        Translate English to Rundi                        & Chrf++ & 18.06\textsuperscript{$\pm$1.16}  & 14.41\textsuperscript{$\pm$2.53} & 20.36\textsuperscript{$\pm$2.88} & \textbf{25.39\textsuperscript{$\pm$1.57}}  & 23.94\textsuperscript{$\pm$3.03} \\
        Translate English to Sesotho                      & Chrf++ & 17.34\textsuperscript{$\pm$3.09}  & 10.2\textsuperscript{$\pm$8.75}  & 19.31\textsuperscript{$\pm$3.94} & 23.85\textsuperscript{$\pm$1.43}  & \textbf{23.9\textsuperscript{$\pm$2.03}}  \\
        Translate English to Swahili                      & Chrf++ & 18.5\textsuperscript{$\pm$0.31}   & 16.28\textsuperscript{$\pm$4.48} & 19.42\textsuperscript{$\pm$2.2}  & 16.16\textsuperscript{$\pm$3.93}  & \textbf{30.6\textsuperscript{$\pm$0.11} } \\
        Translate English to Xhosa                        & Chrf++ & 21.34\textsuperscript{$\pm$2.66}  & 19.96\textsuperscript{$\pm$4.05} & 26.94\textsuperscript{$\pm$1.92} & 15.76\textsuperscript{$\pm$11.49} & \textbf{27.0\textsuperscript{$\pm$1.01}}  \\
        Translate English to Zulu                         & Chrf++ & 21.14\textsuperscript{$\pm$2.6}   & 17.32\textsuperscript{$\pm$3.17} & 28.97\textsuperscript{$\pm$1.14} & 19.29\textsuperscript{$\pm$12.69} & \textbf{40.97\textsuperscript{$\pm$1.10}} \\
        Translate English to Hausa                        & Chrf++ & 25.98\textsuperscript{$\pm$0.27}  & 25.22\textsuperscript{$\pm$0.5}  & 18.28\textsuperscript{$\pm$0.41} & 28.56\textsuperscript{$\pm$0.22}  & \textbf{32.23\textsuperscript{$\pm$0.29}} \\
        Translate English to Igbo                         & Chrf++ & 37.82\textsuperscript{$\pm$0.15}  & 34.8\textsuperscript{$\pm$0.32}  & 20.25\textsuperscript{$\pm$0.68} & 29.89\textsuperscript{$\pm$0.22}  & \textbf{41.87\textsuperscript{$\pm$0.31}} \\
        Translate English to Luganda                      & Chrf++ & 23.15\textsuperscript{$\pm$2.19}  & 20.74\textsuperscript{$\pm$0.36} & 13.43\textsuperscript{$\pm$1.28} & 20.27\textsuperscript{$\pm$2.21}  & \textbf{33.12\textsuperscript{$\pm$0.08}} \\
        Translate English to N. Pidgin              & Chrf++ & 60.57\textsuperscript{$\pm$0.15}  & \textbf{60.12\textsuperscript{$\pm$0.07} }& 23.85\textsuperscript{$\pm$0.64} & 49.72\textsuperscript{$\pm$0.36}  & 59.74\textsuperscript{$\pm$0.18} \\
        Translate English to Swahili                      & Chrf++ & 47.67\textsuperscript{$\pm$3.33}  & 48.95\textsuperscript{$\pm$0.13} & 19.01\textsuperscript{$\pm$1.69} & 40.84\textsuperscript{$\pm$0.31}  & \textbf{53.67\textsuperscript{$\pm$0.15}} \\
        Translate English to Setswana                     & Chrf++ & 29.02\textsuperscript{$\pm$0.35}  & 14.87\textsuperscript{$\pm$0.16} & 11.77\textsuperscript{$\pm$1.61} & 21.25\textsuperscript{$\pm$0.36}  & \textbf{34.05\textsuperscript{$\pm$0.64}} \\
        Translate English to Twi                          & Chrf++ & 21.25\textsuperscript{$\pm$0.22}  & 13.63\textsuperscript{$\pm$1.18} & 11.7\textsuperscript{$\pm$1.13}  & 15.39\textsuperscript{$\pm$3.02}  & \textbf{23.96\textsuperscript{$\pm$0.2}}  \\
        Translate English to Yoruba                       & Chrf++ & 18.41\textsuperscript{$\pm$1.89}  & 15.47\textsuperscript{$\pm$0.4}  & 10.19\textsuperscript{$\pm$0.25} & 18.99\textsuperscript{$\pm$0.27}  & \textbf{24.1\textsuperscript{$\pm$0.06}}  \\
        Translate English to Zulu                         & Chrf++ & 30.99\textsuperscript{$\pm$1.13}  & 13.86\textsuperscript{$\pm$6.85} & 11.34\textsuperscript{$\pm$2.1}  & 10.58\textsuperscript{$\pm$0.77}  & \textbf{34.31\textsuperscript{$\pm$0.2}}  \\
        Translate French to Bambara                       & Chrf++ & 15.75\textsuperscript{$\pm$0.36}  & 6.8\textsuperscript{$\pm$0.97}   & 10.2\textsuperscript{$\pm$1.41}  & 18.28\textsuperscript{$\pm$0.49}  & \textbf{19.65\textsuperscript{$\pm$0.14}} \\
        Translate French to Ghomálá’                      & Chrf++ & 7.0\textsuperscript{$\pm$0.77}    & 5.64\textsuperscript{$\pm$0.44}  & 5.84\textsuperscript{$\pm$3.04}  & \textbf{11.13\textsuperscript{$\pm$0.34}}  & 7.28\textsuperscript{$\pm$2.83}  \\
        Translate French to Ewe                           & Chrf++ & 9.09\textsuperscript{$\pm$2.21}   & 4.75\textsuperscript{$\pm$2.76}  & 6.56\textsuperscript{$\pm$3.19}  & 11.72\textsuperscript{$\pm$1.4}   & \textbf{20.53\textsuperscript{$\pm$0.23}} \\
        Translate French to Fon                           & Chrf++ & 5.24\textsuperscript{$\pm$2.33}   & 5.57\textsuperscript{$\pm$0.63}  & 5.28\textsuperscript{$\pm$1.38}  & 10.94\textsuperscript{$\pm$1.93}  & \textbf{11.76\textsuperscript{$\pm$0.45}} \\
        Translate French to Moore                         & Chrf++ & 7.08\textsuperscript{$\pm$1.33}   & 4.63\textsuperscript{$\pm$2.02}  & 7.18\textsuperscript{$\pm$0.79}  & 10.31\textsuperscript{$\pm$0.64}  & \textbf{11.2\textsuperscript{$\pm$0.54}}  \\
        Translate French to Wolof                         & Chrf++ & 16.27\textsuperscript{$\pm$0.24}  & 2.65\textsuperscript{$\pm$0.11}  & 10.23\textsuperscript{$\pm$1.73} & \textbf{15.73\textsuperscript{$\pm$0.33}}  & 15.58\textsuperscript{$\pm$0.19} \\
        Translate English to N. Pidgin (pidginUNMT) & Chrf++ & 42.12\textsuperscript{$\pm$0.18}  & 37.67\textsuperscript{$\pm$1.64} & 22.53\textsuperscript{$\pm$1.31} & 28.38\textsuperscript{$\pm$0.98}  & \textbf{39.58\textsuperscript{$\pm$0.49}} \\
        Translate Acholi to English                       & Chrf++ & 37.96\textsuperscript{$\pm$0.1}   & 27.18\textsuperscript{$\pm$0.36} & 28.24\textsuperscript{$\pm$0.38} & 31.83\textsuperscript{$\pm$0.07}  & \textbf{41.06\textsuperscript{$\pm$0.06}} \\
        Translate Acholi to Lugbara                       & Chrf++ & 23.41\textsuperscript{$\pm$0.84}  & 19.57\textsuperscript{$\pm$5.04} & 27.18\textsuperscript{$\pm$0.36} & \textbf{31.45\textsuperscript{$\pm$0.29}}  & 30.68\textsuperscript{$\pm$1.02} \\
        Translate Acholi to Luganda                       & Chrf++ & 25.67\textsuperscript{$\pm$0.34}  & 19.59\textsuperscript{$\pm$0.56} & 21.52\textsuperscript{$\pm$0.02} & 28.52\textsuperscript{$\pm$0.63}  & \textbf{33.93\textsuperscript{$\pm$0.48}} \\
        Translate Acholi to Nyankore                      & Chrf++ & 24.02\textsuperscript{$\pm$0.41}  & 17.35\textsuperscript{$\pm$3.35} & 21.38\textsuperscript{$\pm$0.23} & 27.73\textsuperscript{$\pm$0.84}  & \textbf{31.04\textsuperscript{$\pm$0.29}} \\
        Translate Acholi to Ateso                         & Chrf++ & 23.65\textsuperscript{$\pm$1.87}  & 20.07\textsuperscript{$\pm$1.53} & 25.81\textsuperscript{$\pm$0.04} & 31.56\textsuperscript{$\pm$0.57}  & \textbf{31.83\textsuperscript{$\pm$0.99}} \\
        Translate English to Lugbara                      & Chrf++ & 36.83\textsuperscript{$\pm$0.03}  & \textbf{38.3\textsuperscript{$\pm$0.13} } & 37.29\textsuperscript{$\pm$0.12} & 34.3\textsuperscript{$\pm$0.77}   & 35.85\textsuperscript{$\pm$0.01} \\
        Translate English to Luganda                      & Chrf++ & 40.1\textsuperscript{$\pm$0.06}   & 37.56\textsuperscript{$\pm$0.19} & 32.18\textsuperscript{$\pm$1.05} & 38.28\textsuperscript{$\pm$0.2}   & \textbf{45.82\textsuperscript{$\pm$0.04}} \\
        Translate English to Nyankore                     & Chrf++ & 35.93\textsuperscript{$\pm$0.18}  & 34.07\textsuperscript{$\pm$0.12} & 32.59\textsuperscript{$\pm$0.05} & 34.88\textsuperscript{$\pm$0.15}  & \textbf{39.17\textsuperscript{$\pm$0.33}} \\
        Translate English to Ateso (salt)                 & Chrf++ & 37.98\textsuperscript{$\pm$0.11} & 38.93\textsuperscript{$\pm$0.01} & 36.83\textsuperscript{$\pm$1.23} & 37.85\textsuperscript{$\pm$0.4}   & \textbf{39.87\textsuperscript{$\pm$0.47}} \\
        Translate Lugbara to Ateso                        & Chrf++ & 20.55\textsuperscript{$\pm$1.38}  & 18.54\textsuperscript{$\pm$0.65} & 25.6\textsuperscript{$\pm$0.64}  & \textbf{30.48\textsuperscript{$\pm$0.59}}  & 26.43\textsuperscript{$\pm$0.32} \\
        Translate Luganda to Lugbara                      & Chrf++ & 26.79\textsuperscript{$\pm$1.49}  & 23.94\textsuperscript{$\pm$0.48} & 29.13\textsuperscript{$\pm$0.11} & 31.56\textsuperscript{$\pm$0.24}  & \textbf{33.04\textsuperscript{$\pm$0.58}} \\
        Translate Luganda to Ateso                        & Chrf++ & 28.94\textsuperscript{$\pm$0.22}  & 25.11\textsuperscript{$\pm$0.59} & 31.26\textsuperscript{$\pm$0.01} & 33.18\textsuperscript{$\pm$0.05}  & \textbf{35.99\textsuperscript{$\pm$0.45}} \\
        Translate Nyankore to Lugbara                     & Chrf++ & 22.89\textsuperscript{$\pm$0.73}  & 25.75\textsuperscript{$\pm$0.44} & 12.07\textsuperscript{$\pm$0.11} & 30.54\textsuperscript{$\pm$0.2}   & \textbf{31.35\textsuperscript{$\pm$0.2}}  \\
        Translate Nyankore to Luganda                     & Chrf++ & 35.7\textsuperscript{$\pm$0.12}   & 33.73\textsuperscript{$\pm$0.15} & 31.99\textsuperscript{$\pm$0.07} & 35.74\textsuperscript{$\pm$0.54}  & \textbf{41.63\textsuperscript{$\pm$0.0}}  \\
        Translate Nyankore to Ateso                       & Chrf++ & 26.03\textsuperscript{$\pm$0.08}  & 22.35\textsuperscript{$\pm$0.98} & 28.05\textsuperscript{$\pm$0.09} & 30.53\textsuperscript{$\pm$0.13}  & \textbf{32.65\textsuperscript{$\pm$0.62}} \\ \bottomrule
        &\textbf{Average }
        & 25.58	
        & 21.67	
        & 20.50	
        & 25.16	
        & \textbf{31.24} \\ \bottomrule
\end{tabular}
\caption{Performance of various models on MT data using CHRF++}\label{tab:results_chrfplus}
\end{table*}

\section{Linguistic Details}\label{app-sec:linguistic_details}
\noindent{\textbf{Morphology}}
Morphologically, both Hausa and Swahili are classified as agglutinative languages \cite{hausa_2017_jaggar, wals}, characterized by the systematic addition of prefixes, suffixes, and affixes to root words or stems. This process imparts precise grammatical meanings, encompassing tense, case, mood, person, number, and more. Conversely, Yor\`{u}b\'{a} exhibits an analytic structure, relying on word order and discrete function words to denote grammatical relationships, with minimal use of inflections or affixes. The following are examples from the generated (1) Hausa, (2) Swahili, and (3) Yor\`{u}b\'{a}, respectively.

\pex[interpartskip=2ex] 
\a
\begingl
\gla Bai barshi ba  //
\glb neg.masculine leave at-all //
\glft \textit{`he did not leave him'} //
\endgl
\a
\begingl[everygl=\openup.5ex,everygla=,everyglb=, everyglft=\it,aboveglftskip=1.5ex]
\gla   Bata barshi ba  //
\glb Neg.feminine leave at-all //
\glft \textit{`she did not leave him'} //
\endgl
\xe

\pex[interpartskip=2ex] 
\a
\begingl
\gla Ha-ku-mu-a-cha  // 
\glb 3pl.sg.sub-neg-3pl.sg.obj-leave  //
\glft \textit{`He did not leave him'} //
\endgl
\a
\begingl[everygl=\openup.5ex,everygla=,everyglb=, everyglft=\it,aboveglftskip=1.5ex]
\gla   Ha-ku-mu-a-cha  //
\glb 3pl.sg.sub-neg-3pl.sg.obj-leave  //
\glft \textit{`She did not leave him'} //
\endgl
\xe

\pex[interpartskip=2ex] 
\a
\begingl
\gla \`{O}hun \`{o} k\'{u}r\`{o} l\textsubdot{\'{O}}d\textsubdot{\`{O}} \textsubdot{\`{e}}  // 
\glb 3pl.sg.sub neg leave from 3pl.sg.obj //
\glft \textit{`He did not leave him'} //
\endgl
\a
\begingl[everygl=\openup.5ex,everygla=,everyglb=, everyglft=\it,aboveglftskip=1.5ex]
\gla   \`{O}hun \`{o} k\'{u}r\`{o} l\textsubdot{\'{O}}d\textsubdot{\`{O}} \textsubdot{\`{e}}  //
\glb 3pl.sg.sub neg leave from 3pl.sg.obj  //
\glft \textit{`She did not leave him'} //
\endgl
\xe


\noindent{\textbf{Phonology}}
In terms of phonology, Yor\`{u}b\'{a} and Hausa are tonal languages, where pitch distinctions contribute to word differentiation. However, Hausa features a relatively simpler tone system compared to Yor\`{u}b\'{a} and in most cases tone is not marked in Hausa orthography. Only dictionaries and pedagogical materials indicate tone in text. Yor\`{u}b\'{a} on the other hand has three tones and indicating tones in orthography significantly reduces ambiguity \cite{adebara-abdul-mageed-2022-towards}. Swahili, in contrast, is devoid of tones altogether.

\section{Cloze Task Results}
We provide results on the performance of each model on individual languages. We use a dash '-' to indicate that a specific model does not support a language. 

\begin{table}[!]
\centering
\resizebox{0.9\columnwidth}{!}{%
\begin{tabular}{lccccc} \toprule
\textbf{ISO} & \textbf{MT0} & \textbf{MT5} & \textbf{AfriMT5} & \textbf{AfriTeVa} & \textbf{Cheetah} \\ \toprule
afr & 0   & 0   & -       & -        & \textbf{20.45}      \\
amh & 0   & 0   & -       & 0        & 0       \\
bam & -   & -   & 0       & -        & 0       \\
bbj & -   & \colorbox{red!10}{5.21}   & 0       & -        & \textbf{8.45}       \\
ewe & -   & -   & 0       & -        & 0       \\
fon & -   & -   & 0       & -        & 0       \\
hau & 0   & 0   & 0       & 0        & \textbf{13.41}       \\
ibo & 0   & 0   & 0       & 0        & 0      \\
lin & 0   & -   & -       & -        & \textbf{25.35}      \\
lug & -   & -   & 0       & -        & 0       \\
luo & -   & -   & 0       & -        & \textbf{9.35}      \\
mos & -   & -   & 0       & -        & \textbf{14.53}       \\
mlg & 0   & 0   & -       & -        & \textbf{15.65 }     \\
nya & -   & -   & -       & -        & \textbf{7.64}      \\
nyj & &-&-&-&\\
orm & 0   & -   & -       & -        & 0       \\
pcm & -   & -   & 0       & 0        & \textbf{10.10}       \\
sna & 0   & 0   & -       & -        & 0       \\
som & 0   & 0   & -       & 0        & \textbf{10.39}       \\
sot & \colorbox{red!10}{4.69}  & -   & -       & -        & \textbf{15.23}      \\
swa & -   & -   & 0       & 0        & \textbf{7.02}     \\
swh &&&-&-& \\
tir & -   & -  & -       & -        & \textbf{6.33}       \\
tsn & -   & -   & 0       & -        & 0       \\
twi & -   & -   & 0       & -        & 0       \\
wol & -   & -   & 0       & -        & 0       \\
xho & 0   & 0   & -       & -        & \textbf{6.92}       \\
yor & 0   & 3.61   & 0       & 0        & \textbf{6.42}       \\
zul & 0   & 0   & 0       & -       & \textbf{8.05}      \\ \bottomrule
\end{tabular}%
}
\caption{Bleu scores for mask-one cloze task on the union of languages represented in the four models we compare \ourmodelnlg~with. \colorbox{red!10}{Red} describes zero-shot performance greater than $0$.}
\label{tab:cloze_mask_one}
\end{table}

\begin{table}[!]
\centering
\resizebox{0.9\columnwidth}{!}{%
\begin{tabular}{lccccc} \toprule
\textbf{ISO} & \textbf{MT0} & \textbf{MT5} & \textbf{AfriMT5} & \textbf{AfriTeVa} & \textbf{Cheetah} \\ \toprule
afr & 0   & 0   & -       & -        & 0      \\
amh & 0   & 0   & -       & 0        & 0       \\
bam & -   & -   & 0       & -       & 0       \\
bbj & -   & -   & 0       & -        & 0       \\
ewe & -   & -   & 0       & -        & 0       \\
fon & -   & -   & 0       & -        & 0       \\
hau & 0   & -   & 0       & 0        & \textbf{6}       \\
ibo & 0   & -   & 0       & 0        & \textbf{8 }      \\
lin & 0   & -   & -       & -        & 0       \\
lug & -   & -   & 0       & -        & 0       \\
luo & -   & -   & 0       & -        & 0       \\
mos & -   & -   & 0       & -        & 0       \\
mlg & 0   & -   & -       & -        & 0       \\
nya & 0   & 0   & -       & -        & \textbf{12}      \\
nyj & &-&-&-&\\
orm & 0   & -   & 0       & -        & 0       \\
pcm & -   & -   & 0       & 0        & 0       \\
sna & 0   & 0   & -       & -        & 0       \\
som & 0   & 0   & -      & 0        & \textbf{4}       \\
sot & -   & -   & -       & -        & \textbf{10}      \\
swa & -   & -   & 0       & 0        & \textbf{12 }     \\
swh &&&-&-& \\
tir & -   & -   & 0       & 0        & 0       \\
tsn & -   & -   & 0       & -        & 0       \\
twi & -   & -   & 0       & -        & 0       \\
wol & -   & -   & 0       & -        & 0       \\
xho & 0   & 0   & 0       & -        & \textbf{6}       \\
yor & 0   & 0   & 0       & 0        & 0       \\
zul & 0   & 0   & 0       & -        & 0      \\ \bottomrule
\end{tabular}%
}
\caption{Bleu scores for mask-at-least-one cloze task on the union of languages represented in the four models we compare \ourmodelnlg~with.}
\label{tab:cloze_mask_at_least_one}
\end{table}

\section{Annotation}\label{app-sec:annotation}
We gave the following annotation rules to our annotators: Faithfulness refers to how close to the English sentence the model output is. It should be annotated with values between 1 and 5. Faithfulness should be evaluated independently of the fluency of the model output. Below are some detailed explanations for the scale for faithfulness:
\begin{itemize}
    \item Give a value \textbf{1} if model output is not related to the source sentence.
    \item Give a value \textbf{2} if the model output is the opposite of the source sentence.
    \item Give a value \textbf{3} if the model output is somewhat related to the source sentence. It should have some words or phrases that make it related to the source.
    \item Give a value \textbf{4} if the model output is closely related but changes the meaning slightly (e.g difference in gender, number etc)
    \item Give a value \textbf{5} if the model output is an exact translation
    
\end{itemize}

Fluency is how grammatically correct the model is. Faithfulness and fluency should be judged independently. That is, even if the output is not faithful, don't use it to determine the fluency score and vice versa. Here are some detailed explanations on how to assign the values:
\begin{itemize}
    \item Give a value \textbf{1} if model output is completely ungrammatical and nonsensical.
    \item Give a value \textbf{2} if the model output is reasonable but includes some foreign words or gibberish.
    \item Give a value \textbf{3} if the model output contains some grammatical phrases but also contains some ungrammatical phrases. 
    \item Give a value \textbf{4} if the model output is almost grammatical (but may have a few errors like spelling mistakes)
    \item Give a value \textbf{5} if the model output is very fluent and sounds looks like what a native speaker will say. 
    
\end{itemize}

\section{Results on Quality Evaluation}

\begin{figure*}
  \centering
  \begin{subfigure}[b]{0.49\textwidth}
    \includegraphics[width=\linewidth]{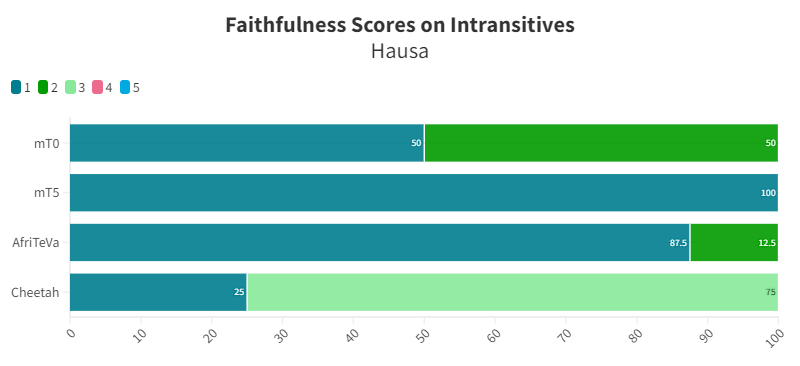}
  \end{subfigure}
  \hfill
   \begin{subfigure}[b]{0.49\textwidth}
    \includegraphics[width=\linewidth]{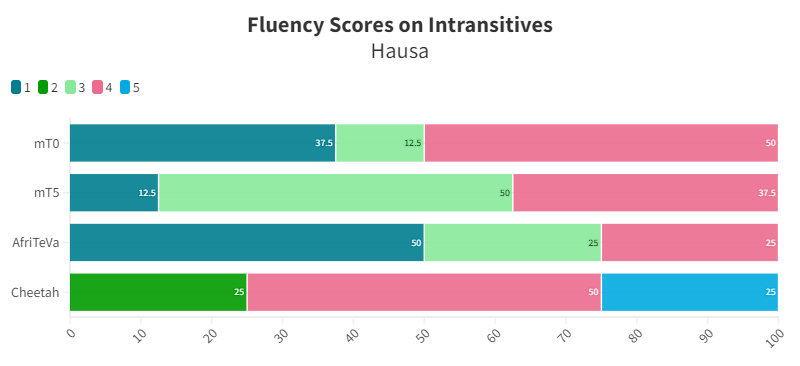}
  \end{subfigure}
  \hfill
   \begin{subfigure}[b]{0.49\textwidth}
    \includegraphics[width=\linewidth]{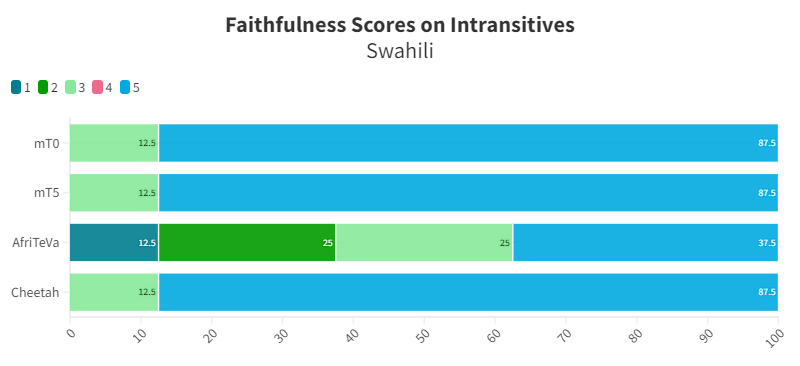}
 \end{subfigure}
 \hfill
  \begin{subfigure}[b]{0.49\textwidth}
    \includegraphics[width=\linewidth]{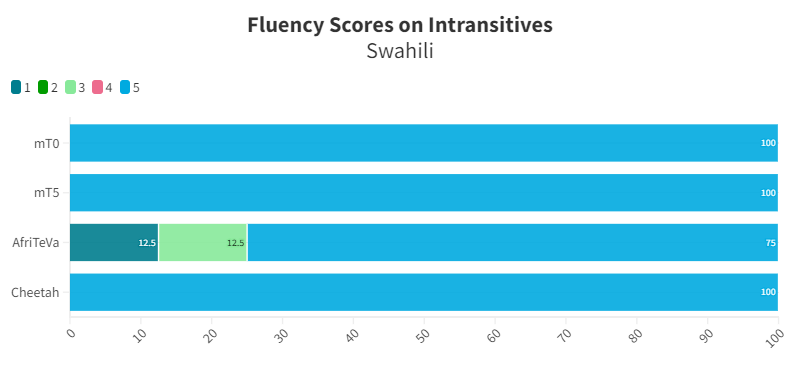}
  \end{subfigure}
 \hfill
  \begin{subfigure}[b]{0.49\textwidth}
    \includegraphics[width=\linewidth]{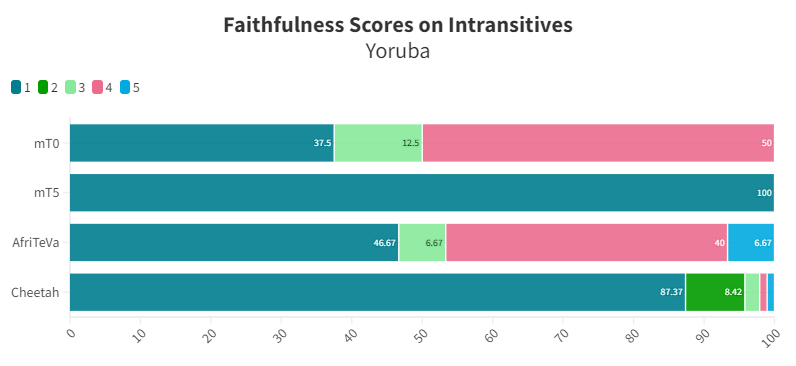}
 \end{subfigure} 
  \hfill
  \begin{subfigure}[b]{0.49\textwidth}
    \includegraphics[width=\linewidth]{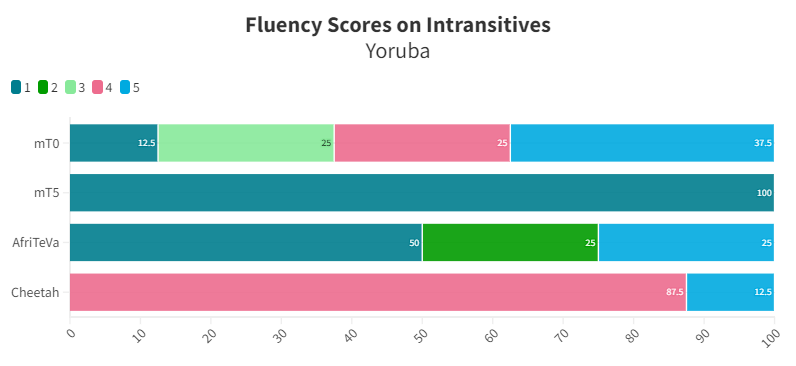}
  \end{subfigure}
  \caption{ Faithfulness and fluency for Intransitives in Hausa, Swahili, and Yor\`{u}b\'{a}}\label{fig:intrans}
\end{figure*}

\begin{figure*}
  \centering
  \begin{subfigure}[b]{0.49\textwidth}
    \includegraphics[width=\linewidth]{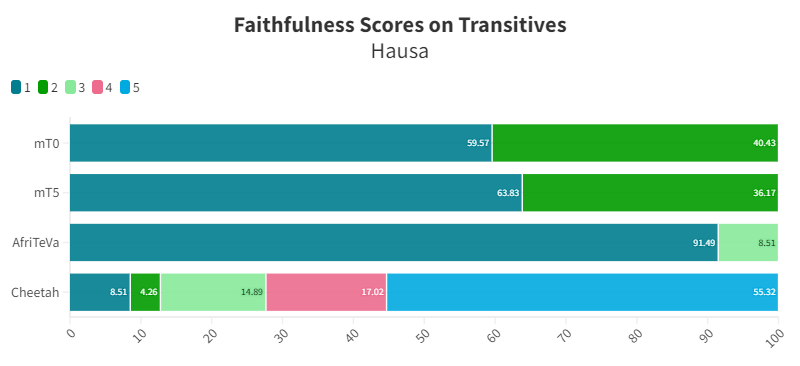}
  \end{subfigure}
  \hfill
   \begin{subfigure}[b]{0.49\textwidth}
    \includegraphics[width=\linewidth]{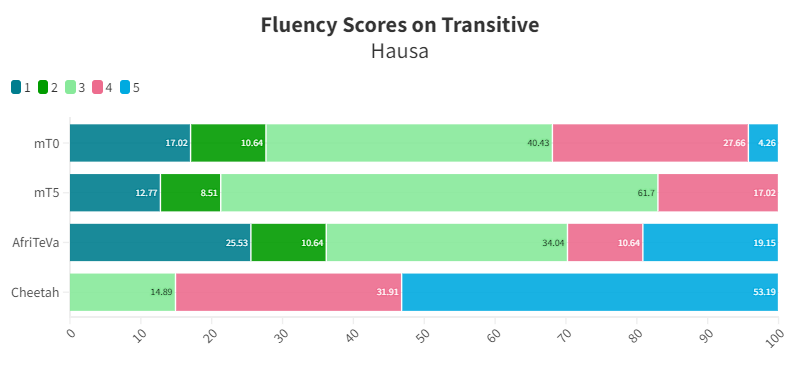}
  \end{subfigure}
  \hfill
   \begin{subfigure}[b]{0.49\textwidth}
    \includegraphics[width=\linewidth]{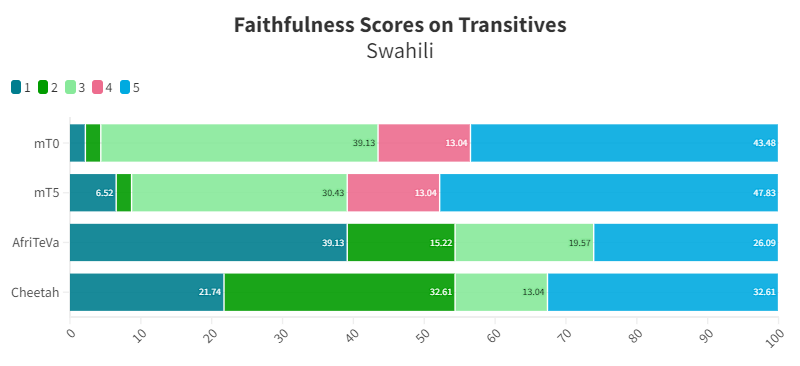}
 \end{subfigure}
 \hfill
  \begin{subfigure}[b]{0.49\textwidth}
    \includegraphics[width=\linewidth]{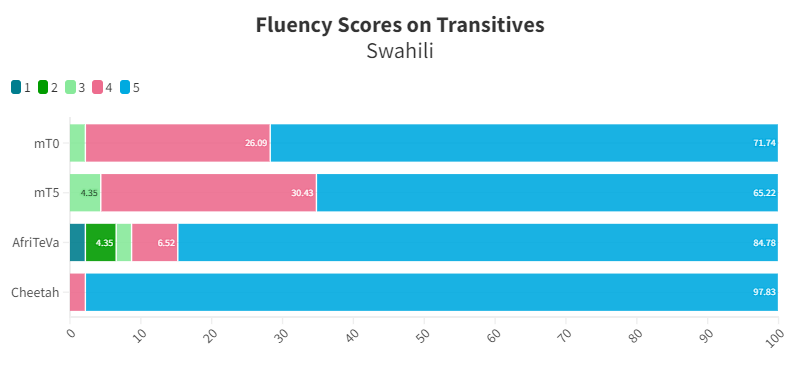}
  \end{subfigure}
 \hfill
  \begin{subfigure}[b]{0.49\textwidth}
    \includegraphics[width=\linewidth]{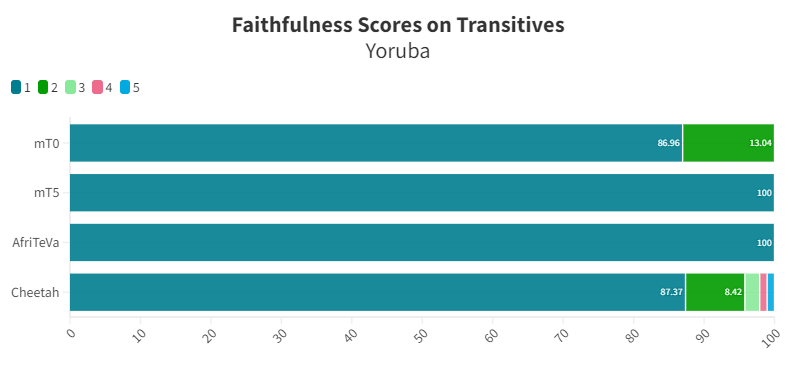}
 \end{subfigure} 
  \hfill
  \begin{subfigure}[b]{0.49\textwidth}
    \includegraphics[width=\linewidth]{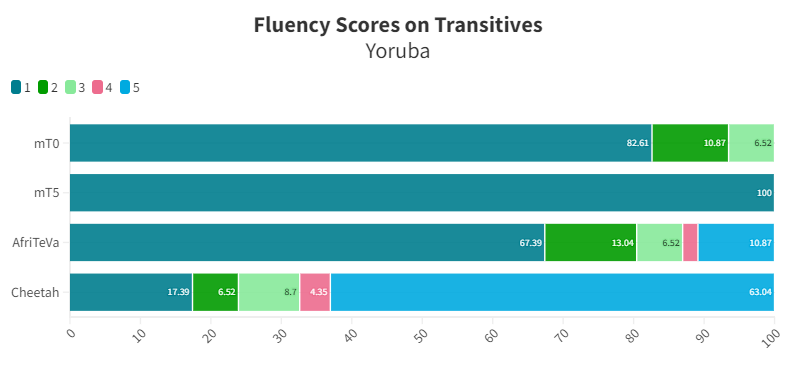}
  \end{subfigure}
  \caption{Faithfulness and fluency for Transitives in Hausa, Swahili, and Yor\`{u}b\'{a}}\label{fig:trans}
\end{figure*}

\begin{figure*}
  \centering
    \includegraphics[width=\linewidth]{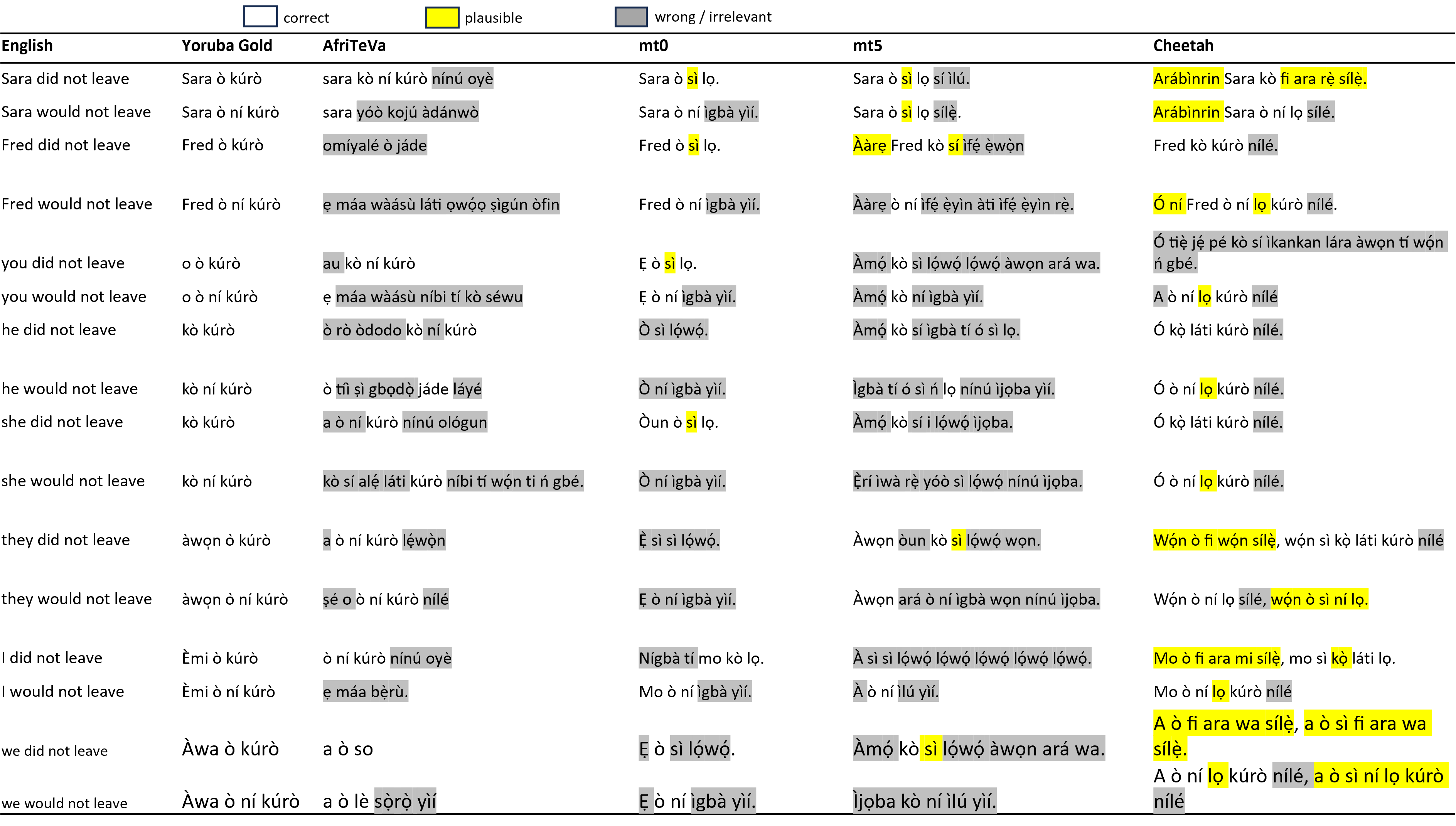}
    \caption{Performance on some intransitive examples in the Yor\`{u}b\'{a} test set. The correct words have no highlights, plausible words or phrases are highlighted with yellow ink while wrong words and phrases are highlighted with grey highlights. We use plausible to refer to words or phrases that can be used in place of the gold or which give additional information.}
  \label{fig:intransitives}
\end{figure*} 

\begin{figure*}
  \centering
  \begin{subfigure}[b]{0.49\textwidth}
    \includegraphics[width=\linewidth]{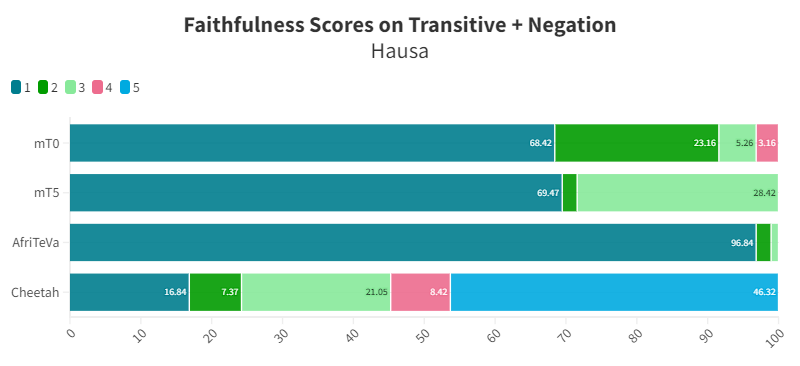}
  \end{subfigure}
  \hfill
   \begin{subfigure}[b]{0.49\textwidth}
    \includegraphics[width=\linewidth]{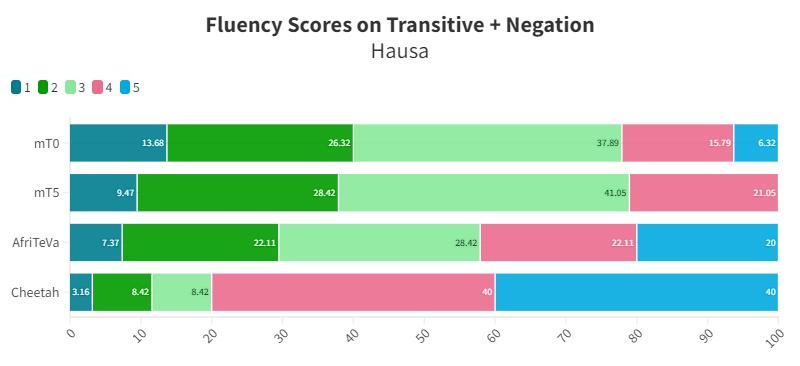}
  \end{subfigure}
  \hfill
   \begin{subfigure}[b]{0.49\textwidth}
    \includegraphics[width=\linewidth]{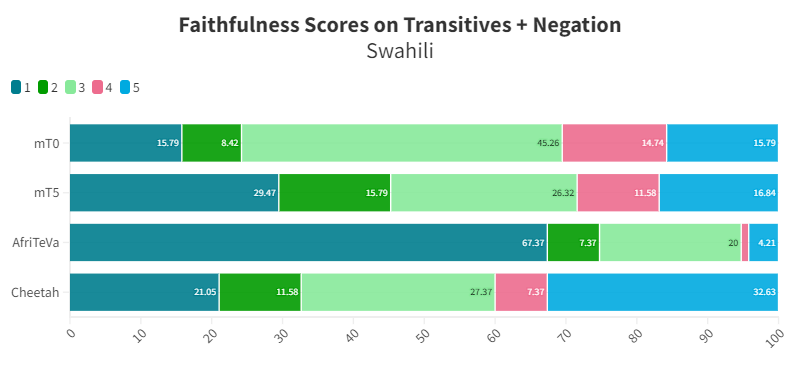}
 \end{subfigure}
 \hfill
  \begin{subfigure}[b]{0.49\textwidth}
    \includegraphics[width=\linewidth]{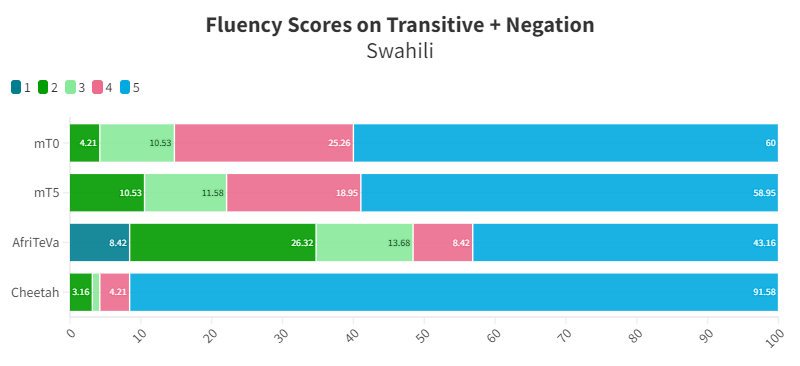}
  \end{subfigure}
 \hfill
  \begin{subfigure}[b]{0.49\textwidth}
    \includegraphics[width=\linewidth]{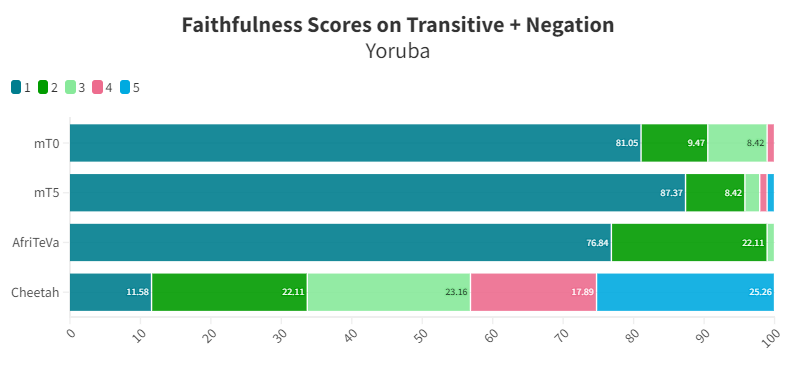}
 \end{subfigure} 
  \hfill
  \begin{subfigure}[b]{0.49\textwidth}
    \includegraphics[width=\linewidth]{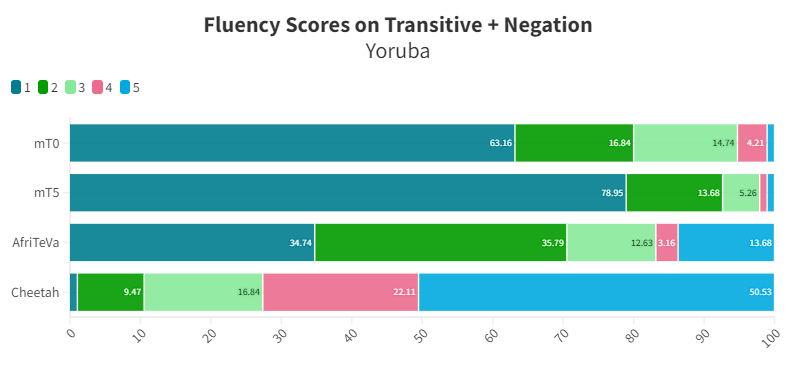}
  \end{subfigure}
  \caption{Faithfulness and fluency for Intransitives + Negation in Hausa, Swahili, and Yor\`{u}b\'{a}}\label{fig:intrans_neg}
\end{figure*}

\begin{figure*}
  \centering
  \begin{subfigure}[b]{0.49\textwidth}
    \includegraphics[width=\linewidth]{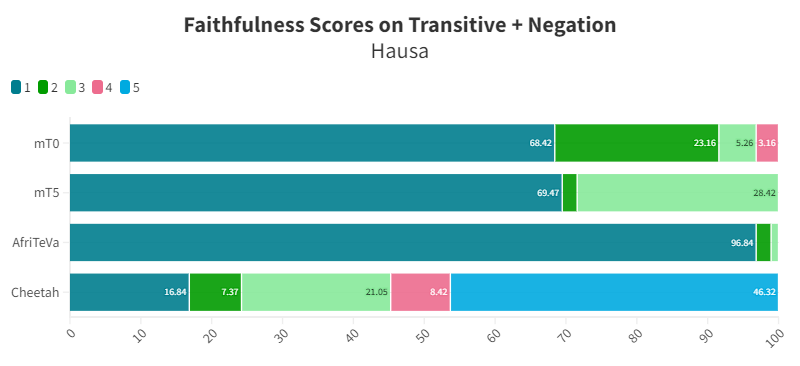}
  \end{subfigure}
  \hfill
   \begin{subfigure}[b]{0.49\textwidth}
    \includegraphics[width=\linewidth]{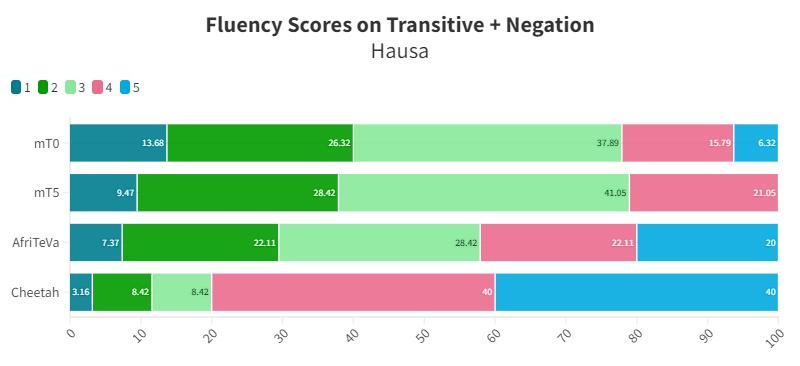}
  \end{subfigure}
  \hfill
   \begin{subfigure}[b]{0.49\textwidth}
    \includegraphics[width=\linewidth]{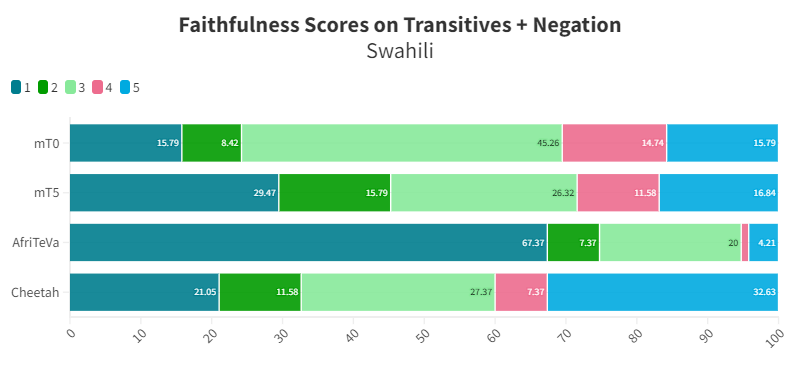}
 \end{subfigure}
 \hfill
  \begin{subfigure}[b]{0.49\textwidth}
    \includegraphics[width=\linewidth]{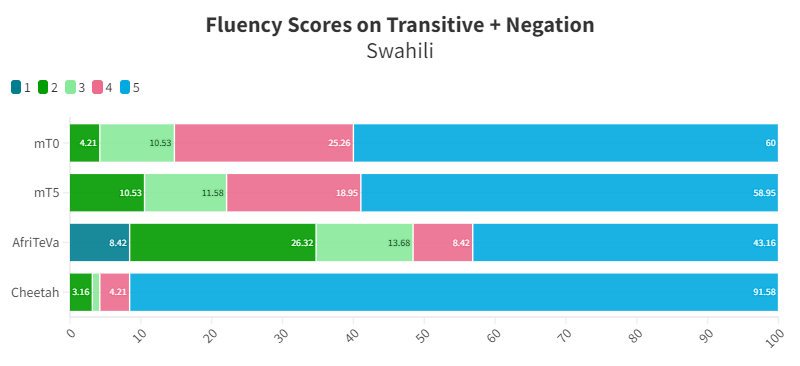}
  \end{subfigure}
 \hfill
  \begin{subfigure}[b]{0.49\textwidth}
    \includegraphics[width=\linewidth]{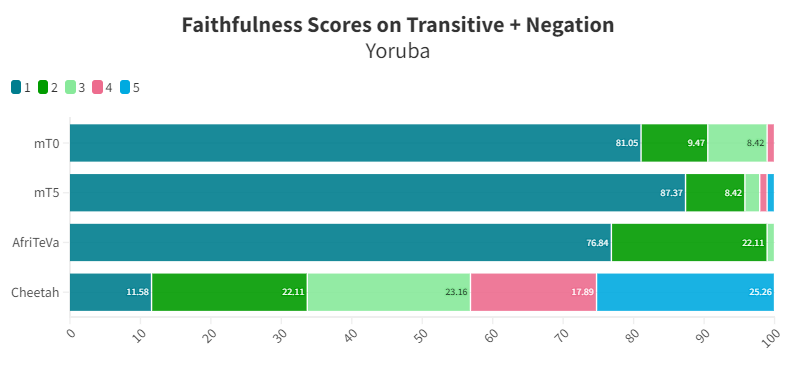}
 \end{subfigure} 
  \hfill
  \begin{subfigure}[b]{0.49\textwidth}
    \includegraphics[width=\linewidth]{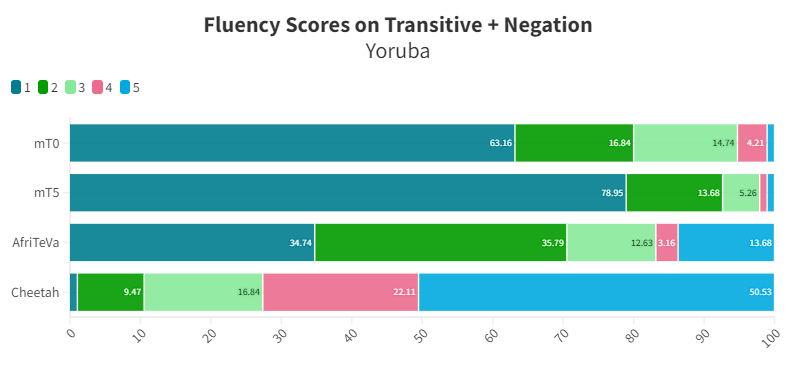}
  \end{subfigure}
  \caption{Faithfulness and fluency for Transitives + Negation in Hausa, Swahili, and Yor\`{u}b\'{a}}\label{fig:trans_neg}
\end{figure*}

\end{document}